\documentclass[10pt,twocolumn,letterpaper]{article}

\usepackage[pagenumbers]{cvpr} 

\usepackage{graphicx}
\usepackage{booktabs}
\usepackage{textpos}
\usepackage{microtype}
\usepackage{marvosym}
\usepackage{amsmath, amsthm, enumitem}
\usepackage[font=small]{caption}
\usepackage{subcaption}
\usepackage{algorithm}
\usepackage{algorithmic}
\usepackage{multirow}
\usepackage{soul}
\usepackage{xcolor}
\usepackage{newfloat}
\usepackage{listings}
\usepackage{pifont}
\usepackage[dvipsnames]{xcolor}
\usepackage{colortbl}
\usepackage{xfrac}
\usepackage{enumitem}
\usepackage{accents}
\usepackage{xspace}
\usepackage{pgfplots}
\usepackage{tikz}
\usepackage{tcolorbox}
\pgfplotsset{compat=1.8}

\newcommand{\cmark}{\textcolor{OliveGreen}{\ding{51}}}
\newcommand{\xmark}{\textcolor{BrickRed}{\ding{55}}}

\newcommand{\model}{\texttt{BiMa}\xspace}
\usepackage{graphicx}
\definecolor{beaublue}{rgb}{0.74, 0.83, 0.9}

\makeatletter
\DeclareRobustCommand\onedot{\futurelet\@let@token\@onedot}
\def\@onedot{\ifx\@let@token.\else.\null\fi\xspace}

\def\eg{\textit{e.g}\onedot} 
\def\ie{\textit{i.e}\onedot}

\makeatother
\definecolor{Gray}{gray}{0.85}
\definecolor{aliceblue}{rgb}{0.94, 0.97, 1.0}
\definecolor{deeppink}{RGB}{255,20,147}
\definecolor{mygray}{gray}{0.95}
\definecolor{ggray}{RGB}{127,127,127}
\definecolor{aliceblue}{rgb}{0.94, 0.97, 1.0}

\definecolor{cvprblue}{rgb}{0.21,0.49,0.74}
\usepackage[pagebackref,breaklinks,colorlinks,allcolors=cvprblue]{hyperref}

\def\eg{\textit{e.g}\onedot} 
\def\ie{\textit{i.e}\onedot}

\makeatother
\definecolor{Gray}{gray}{0.85}
\definecolor{aliceblue}{rgb}{0.94, 0.97, 1.0}
\definecolor{deeppink}{RGB}{255,20,147}
\definecolor{mygray}{gray}{0.95}
\definecolor{ggray}{RGB}{127,127,127}
\definecolor{aliceblue}{rgb}{0.94, 0.97, 1.0}

\title{BiMa: Towards Biases Mitigation for Text-Video Retrieval\\ via Scene Element Guidance}

\author{Huy Le$^{1}$ \quad Nhat Chung$^1$ \quad Tung Kieu$^{2,3}$ \quad Anh Nguyen$^4$ \quad Ngan Le$^{5}$\\
$^1$FPT Software AI Center, Vietnam \quad
$^2$Aalborg University, Denmark \quad
$^3$Pioneer Centre for AI, Denmark \\
$^4$University of Liverpool, UK \quad
$^5$AICV Lab, University of Arkansas, USA \\
{\tt\small \{huylda1, nhatcm3\}@fpt.com} \quad
{\tt\small tungkvt@cs.aau.dk} \quad 
{\tt\small anh.nguyen@liverpool.ac.uk} \quad
{\tt\small thile@uark.edu}}

\begin{document}
\maketitle
\begin{abstract}   
Text-video retrieval (TVR) systems often suffer from visual-linguistic biases present in datasets, which cause pre-trained vision-language models to overlook key details. To address this, we propose \model, a novel framework designed to mitigate biases in both visual and textual representations. Our approach begins by generating scene elements that characterize each video by identifying relevant entities/objects and activities. For visual debiasing, we integrate these scene elements into the video embeddings, enhancing them to emphasize fine-grained and salient details. For textual debiasing, we introduce a mechanism to disentangle text features into content and bias components, enabling the model to focus on meaningful content while separately handling biased information. Extensive experiments and ablation studies across five major TVR benchmarks (\ie, MSR-VTT, MSVD, LSMDC, ActivityNet, and DiDeMo) demonstrate the competitive performance of \model. Additionally, the model's bias mitigation capability is consistently validated by its strong results on out-of-distribution retrieval tasks.
\end{abstract}

\section{Introduction}
\label{sec:intro}

\begin{figure}[t]
\centering
\includegraphics[width=1.\linewidth]{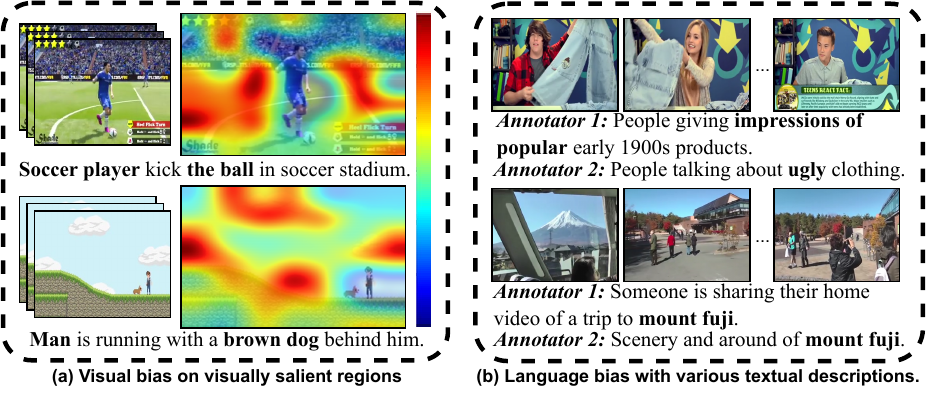}
\vspace{-1.5em}
\caption{\textbf{Illustration of Visual-Linguistic Bias in TVR.} (a) Visual Bias: For each example, video (left of the 1\textsuperscript{st} row), the associated visual feature map (right of the 1\textsuperscript{st} row) extracted by pre-trained \texttt{CLIP}~\cite{li2022clip} and its corresponding Textual query (2\textsuperscript{nd} row). This highlights a bias toward visually dominant regions while overlooking main actors/objects and activities when they occupy smaller regions in the scene. (b) Textual Bias: For each example, video (1\textsuperscript{st} row) and their corresponding textual descriptions (2\textsuperscript{nd} row) from various annotators, capturing emotional responses and personal perspectives.}
\label{fig:teaser}
\vspace{-1.5em}
\end{figure}

The task of retrieving videos based on textual queries and vice versa, known as text-video retrieval (TVR), has rapidly evolved within multimedia information retrieval due to significant advancements in large-scale pre-trained Vision-Language Models (VLMs) such as \texttt{CLIP}~\cite{DBLP:conf/icml/RadfordKHRGASAM21} and \texttt{BLIP}~\cite{DBLP:conf/icml/0001LXH22}. Despite remarkable progress, existing TVR frameworks, including \texttt{TeachCLIP}~\cite{teachclip}, \texttt{TextProxy}~\cite{xiao2025textproxy}, and \texttt{NarVid}~\cite{hur2025narratingthevideo}, largely ignore the underlying visual-linguistic representation biases intrinsic to both the training data and pre-trained VLMs~\cite{liu2024decade,shvetsova2023style,shvetsova2025unbiasing}. Representation biases are systematic deviations in datasets causing models to overemphasize specific features or patterns rather than generalizable and task-relevant aspects~\cite{shvetsova2025unbiasing, liu2024decade}. 
\textit{Liu et al.}~\cite{liu2024decade} demonstrate that models tend to depend excessively on prominent visual concepts and dataset-specific textual patterns, resulting in representations that are tuned to the dataset rather than capturing robust, semantic-rich features. This bias limits the models' ability to generalize to diverse, unseen scenarios. Moreover, the precise nature of the biases learned by neural networks remains largely unclear---some may even contain generalizable and transferable patterns that are not immediately apparent to human observers. \textit{Shvetsova et al.}~\cite{shvetsova2025unbiasing} also confirm that most video datasets are heavily focused on visually salient concepts, such as object bias.
Thus, The absence of bias mitigation within the existing learned TVR frameworks can lead to limitations, resulting in suboptimal performance when exposed to unseen data. We proceed to further describe the bias problem in TVR.

\noindent \textbf{Visual-Linguistic Biases.}
Widely-use TVR datasets such as {MSR-VTT}~\cite{DBLP:conf/cvpr/XuMYR16}, {MSVD}~\cite{DBLP:conf/acl/ChenD11}, {LSMDC}~\cite{lsmdc2016MovieAnnotationRetrieval}, {ActivityNet}~\cite{DBLP:conf/iccv/KrishnaHRFN17}, and {DiDeMo}~\cite{DBLP:conf/iccv/HendricksWSSDR17} often present practical challenges stemming from what we identify as visual-linguistic biases. These biases arise because each dataset is typically created with specific objectives that align with particular research goals, applications, or target users. These biases are variations in visual and textual representations that can skew pre-trained models towards focusing on subjective or dataset-specific features, potentially causing them to overlook essential factual information. The visual-linguistic bias problem can be partitioned into two sub-problems: visual bias and textual bias, as follows.

Visual bias primarily arises due to coarse-grained annotations that often omit critical details such as key actors, objects, or their interactions. Consequently, visual embeddings produced by pre-trained encoders frequently emphasize visually dominant areas, neglecting smaller but semantically crucial components. Figure~\ref{fig:teaser}(a) illustrates this issue, showing how pre-trained \texttt{CLIP} disproportionately focuses on the larger environmental context, marginalizing important yet smaller elements such as the ``man'' and ``brown dog,'' crucial to understanding the scene. 

Textual bias occurs when annotators’ subjective interpretations, cultural perspectives, emotional states, or language usage differences produce varying textual descriptions for identical video scenes. This variability causes models to capture subjective rather than objective semantic content. As depicted in Figure~\ref{fig:teaser}(b), distinct annotators produce emotionally and culturally varied descriptions, demonstrating the need for effective textual bias neutralization.

To address the aforementioned visual-linguistic bias problem, we propose \underline{Bi}as \underline{M}itig\underline{a}tion Text-Video Retrieval (\model). The \model's objectives are twofold: (i) neutralize visual-linguistic biases and (ii) enhance the model's focus on relevant features. To obtain these goals, \model is equipped with three key modules--(i) \textit{Scene Element Construction}, (ii) \textit{Visual Scene Debias}, and (iii) \textit{Textual Content Debias}.
The \textit{Scene Element Construction} module aggregates fine-grained entities and actions, providing structured semantic guidance. Subsequently, the \textit{Visual Scene Debias} module leverages these elements to reorient visual attention toward essential components, thus reducing visual bias. The \textit{Textual Content Debias} module employs a novel disentanglement mechanism to separate textual content from bias-induced variations, ensuring semantic consistency across diverse annotations. Critically, this mechanism is self-supervised, obviating the need for expensive additional annotations.
Finally, to further evaluate our proposed bias mitigation framework, out-of-domain retrieval experiments are leveraged to validate the efficacy and assess the generalization of models beyond training-specific biases by evaluating performance across distinctly different datasets or scenarios, such as training on MSR-VTT and testing on ActivityNet Captions. 
In summary, our contributions are:
\begin{itemize} [noitemsep,topsep=0pt]
    \item Explicitly identifying and defining visual-linguistic biases in TVR task, drawing upon recent bias representation findings~\cite{shvetsova2025unbiasing, liu2024decade}.
    \item Proposing \model, a systematic framework for bias mitigation with three integrated modules: \textit{Scene Element Construction}, \textit{Visual Scene Debias}, and \textit{Textual Content Debias}.
    \item Achieving state-of-the-art performance and robust generalization across multiple TVR benchmarks (MSR-VTT, MSVD, LSMDC, ActivityNet Captions, DiDeMo), supported by comprehensive ablation studies and significant bias reduction in out-of-domain retrieval settings.
\end{itemize}

\section{Problem Definition}
\label{sec:problem}

\begin{figure*}[t]
\centering
\includegraphics[width=\linewidth]{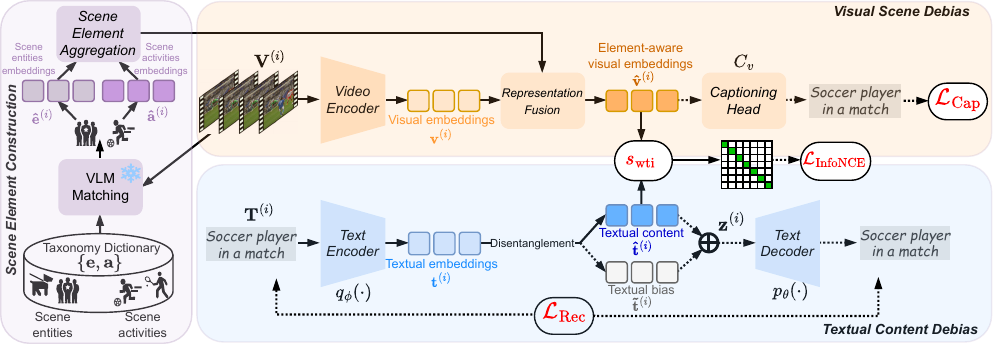}
\caption{\textbf{Overview of \model.} The dashed line represents the model flow used exclusively during training, while the solid line indicates usage in both training and inference. }
\label{fig:framework}
\end{figure*}

\noindent\textbf{Text-Video Retrieval.} Given a text query $\mathbf{T}^{(i)}=\langle\mathbf{T}^{(i)}_1, \mathbf{T}^{(i)}_2, \ldots, \mathbf{T}^{(i)}_{N_{t}}\rangle$ with $N_t$ word tokens and a video $\mathbf{V}^{(i)} = \langle\mathbf{V}^{(i)}_1, \mathbf{V}^{(i)}_2, \ldots, \mathbf{V}^{(i)}_{N_{f}}\rangle$ with $N_f$ frames. 
A Text Encoder encodes $\mathbf{T}^{(i)}$ into a sequence of $N_{t} + 1$ word embeddings ${\mathbf{t}}^{(i)}=\langle{\mathbf{t}}^{(i)}_1, {\mathbf{t}}^{(i)}_2, \ldots, {\mathbf{t}}^{(i)}_{N_{t}}, {\mathbf{t}}^{(i)}_{\texttt{[CLS]}}\rangle$, where $\mathbf{t}_{\texttt{[CLS]}}^{(i)}$ is the global textual embedding. 
A Video Encoder encodes $\mathbf{V}^{(i)}$ into a sequence of $N_f$ frame embedding ${\mathbf{v}}^{(i)}=\langle{\mathbf{v}}^{(i)}_1, {\mathbf{v}}^{(i)}_2, \ldots, {\mathbf{v}}^{(i)}_{N_f}\rangle$.
We aim to learn a cross-modality similarity measure that assigns a high similarity score to  $\mathbf{V}^{(i)}$ and $\mathbf{T}^{(i)}$ and a low similarity score to $\mathbf{V}^{(j)}$ (with $j \neq i$) and $\mathbf{T}^{(i)}$.
Equation~\ref{eq:TVR_problem} formally defines the TVR problem. For brevity, we only focus on the Text-to-Video (\textbf{T2V}). The Video-to-Text (\textbf{V2T}) can be similarly defined.
\begin{equation}
    \max s(\mathbf{V}^{(i)}, \mathbf{T}^{(i)}) = \max \mathbb{P}(\mathbf{V}^{(i)}|\mathbf{T}^{(i)}) = \max \mathbb{P}(\mathbf{v}^{(i)}|\mathbf{t}^{(i)})
    \label{eq:TVR_problem}
\end{equation}

\noindent Here, $s(\cdot,\cdot)$ is a $\mathit{cosine}$ similarity and $\mathbb{P}(\cdot|\cdot)$ indicates the conditional probability.

\vspace{0.5em}
\noindent\textbf{Visual-Linguistic Biases.} When \textit{visual bias} occurs, the visual embedding features of a video $\mathbf{V}^{(i)} $differ from the matching visual embedding features generated for a corresponding query $\mathbf{T}^{(i)}$. We define the visual bias as follows.
\begin{equation}
    \mathbb{P}(\mathbf{v}^{(i)}|\mathbf{V}^{(i)}) \neq \mathbb{P}(\mathbf{v}^{(i)}|\mathbf{T}^{(i)})\\
    \Rightarrow \mathbb{P}(\mathbf{v}^{(i)}|\mathbf{V}^{(i)}) \neq \mathbb{P}(\mathbf{v}^{(i)}|\mathbf{t}^{(i)})
    \label{eq:visual_bias}
\end{equation}
\noindent Equation~\ref{eq:visual_bias} indicates that the Video Encoder does not provide the matching visual embedding features. When a pre-trained model is used as the Video Encoder, the bias can lean towards the pre-trained features.

\noindent When \textit{textual bias} occurs, the probability of the visual embedding features given the textual biases ($\tilde{\mathbf{t}}^{(i)}$) is higher than the visual embedding given the true semantic textual features ($\hat{\mathbf{t}}^{(i)}$). We define the textual bias as follows.
\begin{equation}
           \mathbb{P}(\mathbf{v}^{(i)}|\hat{\mathbf{t}}^{(i)} + \tilde{\mathbf{t}}^{(i)}) > \mathbb{P}(\mathbf{v}^{(i)}|\hat{\mathbf{t}}^{(i)}) \Rightarrow \mathbb{P}(\mathbf{v}^{(i)}|\mathbf{t}^{(i)}) \neq \mathbb{P}(\mathbf{v}^{(i)}|\hat{\mathbf{t}}^{(i)})
    \label{eq:textual_bias}
\end{equation}
\noindent 
Equation~\ref{eq:textual_bias} indicates that the similarity between the visual embedding features and the textual biases is higher than the similarity between the visual embedding features and the true semantic textual features.

\section{Methodology}
\label{sec:methodology}

Figure~\ref{fig:framework} shows the the framework overview. comprising of three main modules: (i) \textit{Scene Element Construction}, (ii) \textit{Visual Scene Debias}, and (iii) \textit{Textual Content Debias}. 
We proceed to introduce these modules in the following sections. 

\subsection{Scene Element Construction}
\label{sec:scene_element_construction} 
\textit{Scene Element Construction} aims to generate scene elements for each video and the corresponding embeddings.
In our proposed framework, scene elements play a central role in debiasing across different modalities.
Scene elements consist of scene entities (noun phrases representing actors or objects) and scene activities (verb phrases representing behaviors or actions), which are drawn from a comprehensive scene taxonomy dictionary (detailed implementation is presented in the \textbf{Appendix}).
Formally, scene elements are represented as a set of nouns and verb phrases $\left\{\mathbf{e},\mathbf{a}\right\}$ that is extracted from large-scale text-video descriptions. 

\vspace{0.5em}
\noindent \textbf{Scene Elements Generation.}
\label{sec:gen_scene} 
Given a video $\mathbf{V}^{(i)}$, we leverage the zero-shot capability of a pre-trained \texttt{CLIP} model with frozen weights to perform text-video matching to identify the most relevant scene elements. 
In particular, we use \texttt{CLIP}'s textual encoder to obtain global textual tokens including a set of $N_e$ scene entities $\mathbf{e} =\left\{\mathbf{e}^{(i)}\right\}_{i=1}^{N_e}$, and a set of $N_a$ scene activities $\mathbf{a} =\left\{\mathbf{a}^{(i)}\right\}_{i=1}^{N_a}$. 
Simultaneously, we feed $\mathbf{V}^{(i)}$ into \texttt{CLIP}'s visual encoder to obtain frame-level visual feature embeddings $\mathbf{v}^{(i)}$. 
The embeddings $\mathbf{v}^{(i)}$ are then temporally aggregated through mean pooling to yield a video-level representation $\bar{\mathbf{v}}^{(i)}$. 
Then, we calculate the similarity between $\bar{\mathbf{v}}^{(i)}$ with scene entities dictionary $\mathbf{e}$ and scene activities dictionary $\mathbf{a}$.
After that, we select the top-$\kappa$ relevant scene entities $\hat{\mathbf{e}}^{(i)} =  \langle \hat{\mathbf{e}}^{(i)}_{1}, \hat{\mathbf{e}}^{(i)}_{2}, \ldots, \hat{\mathbf{e}}^{(i)}_{\kappa} \rangle$ and top-$\kappa$ scene activities $\hat{\mathbf{a}}^{(i)} =  \langle \hat{\mathbf{a}}^{(i)}_{1}, \hat{\mathbf{a}}^{(i)}_{2}, \ldots, \hat{\mathbf{a}}^{(i)}_{\kappa} \rangle$ with the highest similarity scores as ``scene elements'' of $\mathbf{V}^{(i)}$.

\vspace{0.5em}
\noindent  \textbf{Scene Elements Aggregation.}
To enhance the discriminative features corresponding to $\mathbf{V}^{(i)}$, we combine scene entities and scene activities into a unified scene element embedding $\mathbf{c}^{(i)}$. Specifically, we introduce a balancing coefficient $g$ that adjusts the relative importance of scene entities $\hat{\mathbf{e}}^{(i)}$ and scene activities $\hat{\mathbf{a}}^{(i)}$ for each video $\mathbf{V}^{(i)}$. This coefficient $g$ supports a smooth aggregation of embeddings by establishing an association between each feature token and its most relevant counterpart among the other feature tokens. The formulation of $g$ is provided in Equation~\ref{eq:top-k}.
\begin{equation}
\label{eq:top-k}
    g_{(\hat{\mathbf{a}}^{(i)}, \hat{\mathbf{e}}^{(i)})} = 1-\frac{1}{\kappa} \sum^{\kappa}_{l}  \max_j [s(\hat{\mathbf{a}}_l^{(i)}, \hat{\mathbf{e}}_{j}^{(i)})_{j=1}^{\kappa}]
\end{equation}
\noindent Here, $s(\cdot)$ indicates the $\mathit{cosine}$ similarity. The coefficient $g$ is then used to aggregate $\hat{\mathbf{e}}^{(i)}$ and $\hat{\mathbf{a}}^{(i)}$ to obtain scene element embeddings $\mathbf{c}^{(i)}$ as follows.
\begin{equation}
\label{eq:aggregate_embeddding}
    \mathbf{c}^{(i)} = \hat{\mathbf{a}}^{(i)} \oplus g_{(\hat{\mathbf{a}}^{(i)},\hat{\mathbf{e}}^{(i)})} \cdot \hat{\mathbf{e}}^{(i)}
\end{equation} 

\noindent Here, $\oplus$ indicates the element-wise summation operation.

\subsection{Visual Scene Debias}
\label{sec:visual_debias}
\textit{Visual Scene Debias} module aims to mitigate visual biases by explicitly emphasizing fine-grained semantic contents in the visual representation. Particularly, it leverages scene elements to produce element-aware visual scene embeddings. Formally, we aim to mitigate the problem $\mathbb{P}(\mathbf{v}^{(i)}|\mathbf{V}^{(i)}) \neq \mathbb{P}(\mathbf{v}^{(i)}|\mathbf{t}^{(i)})$ (see Equation~\ref{eq:visual_bias}) by augmenting the visual embedding $\mathbf{v}^{(i)}$ with the scene element $\mathbf{c}^{(i)}$ to obtain the augmented visual feature $\hat{\mathbf{v}}^{(i)}$. Then, $\hat{\mathbf{v}}^{(i)}$ is employed as the feature for matching such that $\mathbb{P}(\hat{\mathbf{v}}^{(i)}|\mathbf{V}^{(i)}) \approx \mathbb{P}(\hat{\mathbf{v}}^{(i)}|\mathbf{t}^{(i)})$.
We elaborate on this in the following sections. %

\vspace{0.5em} 
\noindent \textbf{Cross-modality Debias via Representation Fusion.}
Debiasing visual representations involves generating element-aware visual scene embeddings that emphasize relevant information at a lower-dimensional latent space, while capturing interactions between cross-modal features.
More specifically, we aim to align the video embeddings $\mathbf{v}^{(i)}$ and scene element embeddings ${\mathbf{c}}^{(i)}$ to highlight the relevant information $\hat{\mathbf{v}}^{(i)}$ of the video $\mathbf{V}^{(i)}$. 
First, we compute an attention map between a query $\mathbf{q}_{v}$, a key $\mathbf{k}_{c}$, and a value $\mathbf{v}_{c}$. 
Here, $\mathbf{q}_{v}$ represents video embeddings ${\mathbf{v}}^{(i)}$, and both $\mathbf{k}_{c}$ and $\mathbf{v}_{c}$ represent scene element embeddings ${\mathbf{c}}^{(i)}$. 
Then, element-aware visual scene features $\hat{\mathbf{v}}^{(i)}$ are computed through a cross-attention layer~\cite{DBLP:conf/nips/VaswaniSPUJGKP17} as shown in Equation~\ref{eq:ca}.
\begin{equation}
    \begin{aligned}
    \label{eq:ca}
        \mathbf{q}_{v} = \text{Linear}(\mathbf{v}^{(i)}), \mathbf{k}_{c} &= \text{Linear}({\mathbf{c}}^{(i)}),  \mathbf{v}_{c} = \text{Linear}({\mathbf{c}}^{(i)}), \notag \\ 
        \hat{\mathbf{v}}^{(i)} &= \text{softmax}\left(\frac{\mathbf{q}_{v} \mathbf{k}_{c}^{\top}}{\sqrt{d}}\right) \mathbf{v}_{c}. %
    \end{aligned}
\end{equation}
\vspace{0.5em}
\noindent \textbf{Fine-grained Semantic Learning with Captioning Head.}
\label{sec:cap}
Inspired by \texttt{CoCa}~\cite{DBLP:journals/tmlr/YuWVYSW22} that leverages captioning for salient semantic pretraining, we propose to additionally adopt a \textit{Captioning Head} $C_{v}(\cdot)$ parameterized by a stack of $M$ transformer decoders to further learn cross-modal details within element-aware visual scene embeddings $\hat{\mathbf{v}}^{(i)}$. 
In other words, on top of contrastive matching with textual embeddings, the element-aware visual scene embeddings are learned via a text-generative decoding process. 
At each decoding step, $C_{v}(\cdot)$ is trained to predict the subsequent text token $\mathbf{T}^{(i)}_{l}$ with the highest log-likelihood. 
Thus, it learns to auto-regressively maximize the log-likelihood of predicting the input's textual token $\mathbf{T}^{(i)}_{l}$ based on previous tokens $\mathbf{T}^{(i)}_{0:l-1}$. 
As $C_{v}(\cdot)$ relies on $\hat{\mathbf{v}}^{(i)}$, the embeddings $\hat{\mathbf{v}}^{(i)}$ can be established at a deeper alignment of cross-modal representation via Equation~\ref{eq:lm}.
\vspace{-0.7em}
\begin{equation}
\label{eq:lm}
    \mathcal{L}_\mathrm{Cap} = \sum^{N_t}_{l=1} -\log C_v(\mathbf{T}^{(i)}_{l}|\mathbf{T}^{(i)}_{0:l-1}, \hat{\mathbf{v}}^{(i)}).
\end{equation}
\subsection{Textual Content Debias}
\label{sec:textual_debias}

\textit{Textual Content Debias} aims to detach the textual bias from the description and encourage the model to capture the true semantic features via a disentanglement process. 
Formally, we aim to solve the problem $\mathbb{P}(\mathbf{v}^{(i)}|\mathbf{t}^{(i)}) \neq \mathbb{P}(\mathbf{v}^{(i)}|\hat{\mathbf{t}}^{(i)})$ ( Equation~\ref{eq:textual_bias}) by repurposing the \textit{Text Encoder} to decompose every textual description $\mathbf{T}^{(i)}$ into a true semantic component $\hat{\mathbf{t}}^{(i)}$ and a textual bias component $\tilde{\mathbf{t}}^{(i)}$. Then, $\hat{\mathbf{t}}^{(i)}$ is employed as the textual feature for matching with the augmented element-aware visual scene features $\hat{\mathbf{v}}^{(i)}$ which is obtained from the \textit{Visual Scene Debias}, such that $\mathbb{P}(\mathbf{v}^{(i)}|\mathbf{t}^{(i)}) \approx \mathbb{P}(\hat{\mathbf{v}}^{(i)}|\hat{\mathbf{t}}^{(i)})$. We elaborate on this in the below sections.


\vspace{0.5em}
\noindent \textbf{Content-Bias Representation Disentanglement.}
Motivated by the disentanglement ability of $\beta$-\texttt{VAE} framework~\cite{DBLP:conf/iclr/HigginsMPBGBML17}, we propose to use \textit{Text Encoder} to decompose the original textual content $\mathbf{T}^{(i)}$ into two parts in the latent space: a textual content component $\hat{\mathbf{t}}^{(i)}$ and the textual bias component $\tilde{\mathbf{t}}^{(i)}$. These two parts are learned under completely different constraints, thereby separating their representations. 
The former is aligned with the element-aware visual scene features $\hat{\mathbf{v}}^{(i)}$ to enable content matching.
The latter is modeled as a stochastic representation following a Gaussian distribution with the mean ($mu$) and variance (${\sigma}^2$). By doing this, we aim to handle biases as Gaussian noise.
Combining both components will result in the latent representation ${\mathbf{z}}^{(i)} = \hat{\mathbf{t}}^{(i)} + \tilde{\mathbf{t}}^{(i)}$.
Then, the obtained latent representation ${\mathbf{z}}^{(i)}$ is employed to reconstruct the original textual description $\mathbf{T}^{(i)}$.
By reconstructing the two separate components, we ensure the complementary property of each component. In other words, we ensure the combination resulting in ${\mathbf{z}}^{(i)}$ is meaningful.
The reconstruction closely follows the ELBO formulation from conventional \texttt{VAE} framework~\cite{kingma14vae}, which optimizes the variational lower bound on data log-likelihood as shown in Equation~\ref{eq:ELBO}.
\begin{equation}
    \begin{aligned}
        \label{eq:ELBO}
    \mathcal{L}_{\mathrm{VAE}} = &\underbrace{-\mathbb{E}_{\mathbf{z}^{(i)}\sim {q}_{\phi}(\mathbf{z}^{(i)}|\mathbf{T}^{(i)})}\left[\operatorname{log}{p}_{\theta}(\mathbf{T}^{(i)}|\mathbf{z}^{(i)}) \right]}_{\scalebox{1.1}{\quad\quad\quad$\mathcal{L}_{\mathrm{Rec}}$}} \\ 
    &+ \underbrace{D_{\mathbb{KL}}\left({q}_{\phi}(\mathbf{z}^{(i)}|\mathbf{T}^{(i)})||p_{\theta}(\mathbf{z}^{(i)})\right)}_{\scalebox{1.1}{\quad\quad$\mathcal{L}_{\mathbb{KL}}$}}
    \end{aligned}
\end{equation}
\noindent Here, $p_{\theta}(\mathbf{z}^{(i)})$ is a prior modeled as $\mathcal{N}(\mathbf{0},\mathbf{I})$. The conditional probability distributions ${q}_{\phi}(\mathbf{z}^{(i)}|\mathbf{T}^{(i)}), {p}_{\theta}(\mathbf{T}^{(i)}|\mathbf{z}^{(i)})$ are estimated by the \textit{Text Encoder} and \textit{Text Decoder} networks, respectively. Here, $\mathbf{z}^{(i)}$ plays the role of textual embedding, i.e., $\mathbf{z}^{(i)} = \mathbf{t}^{(i)}$. Next, $D_{\mathbb{KL}}(\cdot,\cdot)$ is the Kullback-Leibler (KL) divergence
aiming to enforce the posterior and the prior distribution to be close to each other. The KL divergence plays as a disentanglement factor to enforce the disentanglement process of the bias from the original textual features. Also, it serves as a regularizer that prevents the latent representation from collapsing to zero.
The reconstruction term $\mathcal{L}_{\mathrm{Rec}}$ in the \texttt{VAE} loss measures the log-likelihood of \textit{Text Decoder} ${p}_{\theta}$ to auto-regressively reconstruct the ground-truth input's textual description $\mathbf{T}^{(i)}$ as defined in Equation~\ref{eq:rec}.
\begin{equation}
    \label{eq:rec}
        \mathcal{L}_{\mathrm{Rec}} = \sum^{N_t}_{l=1} -\log {p}_{\theta}(\mathbf{T}^{(i)}_{l}|\mathbf{T}^{(i)}_{0:l-1}, {\mathbf{z}}^{(i)}).
\end{equation}

To capture the unpredictable variations of the textual biases, we model $\tilde{\mathbf{t}}^{(i)}$ as a probabilistic representation. While these features are not explicitly matched to visual representations, they act as distractors for TVR and thus must be modeled to facilitate effective disentanglement. Following this insight, we model the latent representation $\tilde{\mathbf{t}}^{(i)}$ as a multivariate Gaussian distribution.
\begin{equation}
        q_{\phi}(\tilde{\mathbf{t}}^{(i)}|\mathbf{T}^{(i)}) \sim \mathcal{N} ({\mu}, \mathrm{diag}({\sigma}^2))
    \label{eq:distribution}
\end{equation}
To ensure stable training, we apply the reparameterization trick as formulated in Equation~\ref{eq:reparameterization}.

\begin{equation}
    q_{\phi}(\tilde{\mathbf{t}}^{(i)}|\mathbf{T}^{(i)}) = \mu + \sigma
    \cdot \epsilon,
\label{eq:reparameterization}
\end{equation}

\noindent where $\epsilon$ is an auxiliary noise variables and $\epsilon \sim \mathcal{N}(\mathbf{0},\mathbf{I})$.

After modeling the latent representation of textual biases as a distribution $q_{\phi}(\tilde{\mathbf{t}}^{(i)}|\mathbf{T}^{(i)})$, we generate textual bias embedding $\tilde{\mathbf{t}}^{(i)}$ by sampling $K$ instances from this distribution during the training process as shown in Equation~\ref{eq:sampling-k}. This process allows for gradient propagation through the sampled embeddings of bias modeling. 

\begin{equation}
    \tilde{\mathbf{t}}^{(i)}=\langle\tilde{\mathbf{t}}^{(i)}_{1}, \cdots, \tilde{\mathbf{t}}^{(i)}_{2}, \tilde{\mathbf{t}}^{(i)}_{K}\rangle \sim q_{\phi}(\tilde{\mathbf{t}}^{(i)}|\mathbf{T}^{(i)}) 
    \label{eq:sampling-k}
\end{equation}

\vspace{0.5em}
\noindent \textbf{Element-Aware Content Learning by Alignment.} 
Our framework performs disentanglement between content and bias representations by explicitly matching the content embeddings with element-aware visual embeddings, and performing reconstruction of textual disentanglement. In particular, our model is trained via contrastive alignment between element-aware visual embeddings $\hat{\mathbf{v}}^{(i)}$ and textual content embeddings $\hat{\mathbf{t}}^{(i)}$ to optimize their cross-modal similarity (see Equation~\ref{eq:infonce}). Meanwhile, the original texts ${\mathbf{t}}^{(i)}$ are disentangled as content $\hat{\mathbf{t}}^{(i)}$ and bias $\tilde{\mathbf{t}}^{(i)}$, and reconstructed via ${\mathbf{z}}^{(i)}$ and $\mathcal{L}_{\mathrm{VAE}}$. Thus, by the process of elimination, we capture bias representations $\tilde{\mathbf{t}}^{(i)}$ as residual information that is less relevant to element-aware visual embeddings $\hat{\mathbf{v}}^{(i)}$, yet are part of the original information ${\mathbf{t}}^{(i)}$.

\subsection{Objective Function}
\label{sec:total_loss}

To enhance the model's ability to discern relevant features within the data, we replace mean pooling with an advanced weighted pooling mechanism termed Weighted Token Interaction (WTI)~\cite{DRLTVR2022} in order to learn the token-level interaction between cross-modal embeddings. The token-level matching function is denoted as $s_{\mathrm{wti}}(\cdot)$.

For contrastive learning, we adopt the $\mathrm{InfoNCE}$ loss function~\cite{DBLP:journals/corr/abs-1807-03748} to optimize cross-modal similarity between $\hat{\mathbf{v}}^{(i)}$ and $\hat{\mathbf{t}}^{(j)}$ of two sample $i$ and $j$ as shown in Equation~\ref{eq:infonce}.
\begin{equation}
    \begin{aligned}
        \label{eq:infonce}
        \mathcal{L}_{\mathrm{InfoNCE}}=-\frac{1}{2B}\sum_{i=1}^{B}\log\frac{\exp(s_{\mathrm{wti}}({\hat{\mathbf{v}}^{(i)},\hat{\mathbf{t}}^{(i)}})\cdot \tau)}{\sum_{j=1}^{B}\exp(s_{\mathrm{wti}}({\hat{\mathbf{v}}^{(j)},\hat{\mathbf{t}}^{(i)}})\cdot \tau)} \\
        -\frac{1}{2B}\sum_{i=1}^{B}\log\frac{\exp(s_{\mathrm{wti}}({\hat{\mathbf{v}}^{(i)}, \hat{\mathbf{t}}^{(i)}})\cdot \tau)}{\sum_{j=1}^{B}\exp(s_{\mathrm{wti}}({\hat{\mathbf{v}}^{(i)}, \hat{\mathbf{t}}^{(j)}})\cdot \tau)},
    \end{aligned}
\end{equation}

\noindent where $\tau$ is a learnable scaling factor and $B$ is the batch size.

Our process to mitigate visual-linguistic biases exhibits during both training and inference by our end-to-end architecture. However, bias modeling is not required during the inference. The overall objective is defined in Equation~\ref{eq:total_loss}.
\begin{equation}
    \mathcal{L} = \mathcal{L}_{\mathrm{InfoNCE}} + 
    \lambda_{\mathrm{Cap}} \cdot \mathcal{L}_{\mathrm{Cap}} + 
    \lambda_{\mathrm{Rec}} \cdot \mathcal{L}_{\mathrm{Rec}} + \lambda_{\mathbb{KL}} \cdot \mathcal{L}_{\mathbb{KL}}
    \label{eq:total_loss}
\end{equation}

\noindent Here, $\lambda_{\mathtt{cap}}$, $\lambda_{\mathtt{rec}}$, and $\lambda_{\mathbb{KL}}$ are the hyperparameters that control the trade-off among three loss terms.
\section{Experiments}
\begin{table*}[!t]
    \centering
    \footnotesize
    \caption{\textbf{T2V} and \textbf{V2T} performance comparisons on {MSR-VTT} dataset. The best and second best are \textbf{bold} and \underline{underlined}. Two-stage methods are marked with $\dagger$.}
    \vspace{-0.7em}
    \label{table:msrvtt}
 \setlength{\tabcolsep}{12pt}
    \resizebox{1.0\linewidth}{!}{%
    \begin{tabular}{l|l|cccc|cccc}
    \toprule
    \multirow{2}{*}{\textbf{PT}} & \multicolumn{1}{c|}{\multirow{2}{*}{\textbf{Methods}}} & 
    \multicolumn{4}{c|}{\textbf{\textbf{T2V}}} & \multicolumn{4}{c}{\textbf{\textbf{V2T}}}\\
    
     & & R@1$\uparrow$ & R@5$\uparrow$ & R@10$\uparrow$ & Rsum$\uparrow$ & R@1$\uparrow$ & R@5$\uparrow$ & R@10$\uparrow$ & Rsum$\uparrow$  \\
    
    \midrule
    
    \multirow{6}{*}{\cmark} & {\texttt{ClipBERT}~\cite{DBLP:conf/cvpr/LeiLZGBB021}}~{CVPR'20}  & 22.0 & 46.8 & 59.9 & 128.7  & - & - & - & - \\
    & {\texttt{SupportSet}~\cite{DBLP:conf/iclr/Patrick0AMHHV21}}~{ICLR'21} & 30.1 & 58.5 & 69.3 & 157.9 & 28.5 & 58.6 & 71.6 & 158.7 \\
    & {\texttt{Frozen}~\cite{DBLP:conf/iccv/BainNVZ21}}~{ICCV'21} & 32.5 & 61.5 & 71.2 & 165.2 & - & - & - & -  \\
    
    & 
    {\texttt{TMVM}~\cite{DBLP:conf/nips/LinW0ZG0S22}}~{NeurIPS'22} &  36.2 & 64.2 & 75.7 & 176.1 & 34.8 & 63.8 & 73.7 & 172.3\\

    & {\texttt{RegionLearner}~\cite{DBLP:conf/aaai/YanSGW0C023}}~{AAAI'23}  & 36.3 & 63.9 & 72.5 & 172.7 & 34.3 & 63.5 & 73.2 & 171.0 \\
    
    &
    {\texttt{In-Style}~\cite{shvetsova2023style}}~{ICCV'23} & 36.2 & 61.8 & 71.9 & 169.9 & - & - & -  & - \\

    
    \midrule
    
    & {\texttt{CenterCLIP}~\cite{DBLP:conf/sigir/ZhaoZWY22}}~{SIGIR'22} & 44.2 & 71.6 & 82.1 & 197.9  & 42.8 & 71.7 & 82.2 & 196.7 \\
    
    \multirow{18}{*}{\xmark} & {\texttt{CLIP4Clip}~\cite{DBLP:journals/ijon/LuoJZCLDL22}}~{Neurocomputing'22}  & 44.5 & 71.4 & 81.6 & 197.5  & 42.7 & 70.9 & 80.6 & 194.2 \\
    
    & {\texttt{EMCL-Net}~\cite{DBLP:conf/nips/JinHLWGSC022}}~{NeurIPS'22} &  46.8 & 73.1 & 83.1 & 203.0  & 46.5 & 73.5 & 83.5 & 203.5 \\
    
    & {\texttt{X-Pool}~\cite{DBLP:conf/cvpr/GortiVMGVGY22}}~{CVPR'22} & 46.9 & 72.8 & 82.2 & 201.9 & - & - & - & - \\
     
    & {\texttt{TS2-Net}~\cite{DBLP:conf/eccv/LiuXXCJ22}}~{ECCV'22} &  47.0 & 74.5 & {83.8} & 205.3  & 45.3 & 74.1 & 83.7 & 203.1 \\
    
    
    & {\texttt{HBI}~\cite{jin2023video}}~{CVPR'23} {$\dagger$}   & 48.6 & 74.6 & 83.4 & 206.6  & 46.8 & 74.3 & 84.3 & 205.4 \\

    & {\texttt{DiCoSA}~\cite{ijcai2023p0104}}~{IJCAI'23}   & 47.5 & 74.7 & {83.8} & 206.0  & 46.7 & {75.2} & 84.3 & {206.2} \\
    
    & {\texttt{UATVR}~\cite{Fang_2023_ICCV}}~{ICCV'23} &  47.5 & 73.9 & 83.5 & 204.9  & 46.9 & 73.8 & 83.8 & 204.5 \\
    
    & {\texttt{DiffusionRet}~\cite{jin2023diffusionret}}~{ICCV'23} {$\dagger$}  & {49.0} & {75.2} & 82.7 & {206.9} & {47.7} & 73.8 & {84.5} & 206.0 \\  
    
    &{\texttt{PAU}~\cite{jin2023diffusionret}}~{NeurIPS'23} & 48.5 & 72.7 & 82.5 & 203.7 & 48.3 & 73.0 & 83.2 & 204.5 \\
     
    & {\texttt{DGL}~\cite{yang2024dgl}}~{AAAI'24} & 45.8 & 69.3 & 79.4 & 194.5 & - & - & - & - \\  
    
    & {\texttt{EERCF}~\cite{DBLP:conf/aaai/ChenWLQMS23}}~{AAAI'24} & 47.8 & 74.1 & 84.1 & 206.0 & 44.7 & 74.2 & 83.9 & 202.8 \\
    
    & {\texttt{TeachCLIP}~\cite{teachclip}}~{CVPR'24} & 46.8 & 74.3 & 82.6 & 203.7 & - & - & - & - \\  
    
    & {\texttt{DITS}~\cite{DBLP:conf/nips/WangWLGDRRT24}}~{NeurIPS'24} & 51.9 & 75.7 & 84.6 & 212.2 & - & - & - & - \\  
    
    & {\texttt{TextProxy}~\cite{xiao2025textproxy}}~{AAAI'25} & \underline{52.3} & \underline{77.8} & \underline{85.8} & \underline{215.9} & - & - & - & - \\  
    
    & {\texttt{TempMe}~\cite{Leqi2025iclr}}~{ICLR'25} & 46.1 & 71.8 & 80.7 & 198.6 & 45.6 & 72.4 & 81.2 & 199.2 \\  

    & {\texttt{NarVid}~\cite{hur2025narratingthevideo}}~{CVPR'25} & 51.0 & 76.4 & 85.2 & 212.6 & \underline{50.0} & \underline{75.4} & \underline{83.8} & \underline{209.2} \\

    \rowcolor{beaublue!50} & \textbf{\texttt{\model} ({ours})} & \textbf{53.5} & \textbf{78.6} & \textbf{86.5} & \textbf{218.6}  & \textbf{52.2} & \textbf{77.1} & \textbf{85.3} & \textbf{214.6} \\

\bottomrule	
\end{tabular}
}
\vspace{-0.5em}
\end{table*}

\begin{table*}[t]
\footnotesize
\caption{\textbf{T2V} comparisons on {MSVD}, {LSMDC}, {ActivityNet}, and {DiDeMo} datasets. The best and second best are \textbf{bold} and \underline{underlined}.}
\vspace{-0.7em}
\setlength{\tabcolsep}{3.0pt}
\renewcommand{\arraystretch}{0.9}
\resizebox{1.0\linewidth}{!}{
\begin{tabular}{l|cccc|cccc|cccc|cccc}
\toprule
\multicolumn{1}{c|}{\multirow{2}{*}{\textbf{Methods}}} & 
\multicolumn{4}{c|}{\textbf{MSVD}} & 
\multicolumn{4}{c|}{\textbf{LSMDC}} & \multicolumn{4}{c|}{\textbf{ActivityNet}} & \multicolumn{4}{c}{\textbf{DiDeMo}} \\ 
 & 
{R@1$\uparrow$} & {R@5$\uparrow$} & {R@10$\uparrow$} & {Rsum$\uparrow$} & {R@1$\uparrow$} & {R@5$\uparrow$} & {R@10$\uparrow$} & {Rsum$\uparrow$} & {R@1$\uparrow$} & {R@$\uparrow$↑} & {R@10$\uparrow$} & {Rsum$\uparrow$} & {R@1$\uparrow$} & {R@5$\uparrow$} & {R@10$\uparrow$} & {Rsum$\uparrow$} \\ 
\midrule 

{\texttt{ClipBERT}} & - & - & - & {-} & - & - & - & {-} & 21.3 & 49.0 & 63.5  & {133.8} & 20.4 & 48.0 & 60.8 & {129.2} \\

{\texttt{SupportSet}} & 28.4 & 60.0 & 72.9 & {161.3} & - & - & - & {-} & 29.2 & 61.6 & 94.7 & {185.5} & - & - & - & {-} \\

{\texttt{Frozen}} & 33.7 & 64.7 & 76.3 & {174.7} & 15.0 & 30.8 & 40.3 & {86.1} & - & - & - & {-} & 34.6 & 65.0 & 74.7 & {174.3} \\

{\texttt{TMVM}} & 36.7 & 67.4 & 81.3 & {185.4} & 17.8 & 37.1 & 45.9 & {100.8 } & - & - & - & {-} & 36.5 & 64.9 & 75.4 & {176.8} \\

{\texttt{RegionLearner}} & 44.0 & 74.9 & 84.3 & {203.2} & 17.1 & 32.5 & 41.5 & {91.1} & - & - & - & {-} & 32.5 & 60.8 & 72.3 & {165.6} \\

{\texttt{In-Style}} & 44.8 & 72.5 & 81.2 & {198.5} & 16.1 & 33.6 & 39.7  & {89.4} &  - & - & - & {-} & 32.1 & 61.9 & 71.2 & {{165.2}} \\

{\texttt{CenterCLIP}} & {45.2} & {75.5} & {84.3} & {205.0} & 22.6 & 41.0 & {49.1} & {{112.7}} & {40.5} & {72.4} & {83.6} & {{196.5}} & {42.8} & {68.5} & {79.2} & {{190.5}} \\

{\texttt{CLIP4Clip}} & {45.2} & {75.5} & {84.3} & {205.0} & 22.6 & 41.0 & {49.1} & {{112.7}} & {40.5} & {72.4} & {83.6} & {{196.5}} & {42.8} & {68.5} & {79.2} & {{190.5}} \\

{\texttt{EMCL-Net}} & 42.1 & 71.3 & 81.1  & {194.5} & 23.9 & 42.4 & 50.9 & {117.2} & 41.2 & 72.7 & 83.6  & {197.5} & 45.3 & 74.2 & 82.3  & {201.8} \\

{\texttt{X-Pool}} & 47.2 & 77.4 & {86.0}  & {{210.6}} & 25.2 & 43.7 & 53.5  & {122.4} & - & - & - & {-} & - & - & - & {-} \\

{\texttt{TS2-Net}} & - & - & - & {-} & 23.4 & 42.3 & 50.9 & {116.6} & 41.0 & 73.6 & 84.5  & {199.1} & 41.8 & 71.6 & 82.0  & {195.4} \\

{\texttt{HBI}} & - & - & - & {-} & - & - & - & {-} & 42.2 & 73.0 & 84.6  & {199.8} & {46.9} & 74.9 & 82.7  & {204.5} \\

{\texttt{DiCoSa}} & {47.4} & 76.8 & {86.0} & {210.2} & {25.4} & {43.6} & 54.0 & {{123.0}} & 42.1 & 73.6 & 84.6 & {200.3} & 45.7 & 74.6 & 83.5 & {203.8} \\

{\texttt{UATVR}} & 46.0 & 76.3 & 85.1  & {207.4} &  - & - & - & {-} & - & - & - & {-} & 43.1 & 71.8 & 82.3  & {197.2} \\

{\texttt{DiffusionRet}} & 46.6 & 75.9 & 84.1 & {206.6} & 24.4 & 43.1 & {54.3} & {121.8} & {45.8} & {75.6} & {86.3} & {{207.7}} & 46.7 & 74.7 & 82.7 & {204.1} \\

{\texttt{PAU}} & 47.3 & 77.4 & 85.5  & {210.2} &  - & - & - & {-} &  - & - & - & {-} & {48.6} & {76.0} & {84.5} & {{209.1}} \\

{\texttt{EERCF}} & 47.0 & {77.5} & 85.4 & {209.9} & - & - & - & {-} & 43.1 & 74.5 & 86.0  & {203.6} & - & - & - & {-} \\

{\texttt{DGL}} & - & - & - & - & 21.4 & 39.4 & 48.4 & 109.2 & 38.6 & 69.2 & 81.6 & 189.4 & - & - & - & - \\

{\texttt{TeachCLIP}} & {47.4} & 77.3 & 85.5  & {210.2} & - & - & - & {-} & 42.2 & 72.7 & 85.2  & {200.1} & 43.7 & 71.2 & 81.1  & {196.0} \\

{\texttt{DITS}} & - & - & - & - & \underline{28.2} & \underline{47.3} & \underline{56.6} & \underline{132.1} & - & - & - & - & {51.1} & {77.9} & 85.8 & {214.8} \\

{\texttt{TextProxy}} & - & - & - & - & - & - & - & - & \underline{53.0} & \underline{80.9} & \underline{89.6} & \underline{223.5} & {50.6} & {76.9} & {86.0} & {213.5} \\

{\texttt{TempMe}} & - & - & - & - & 23.5 & 41.7 & 51.8 & 117.0 & 44.9 & 75.2 & 85.5 & 205.6 & 48.0 & 72.4 & 81.8 & 202.2 \\

{\texttt{NarVid}} & \underline{53.1} & \underline{81.4} & \underline{88.8} & \underline{223.3} & - & - & - & - & - & - & - & - & \underline{53.4} & \underline{79.1} & \underline{86.3} & \underline{218.8} \\

\rowcolor{beaublue!50} \textbf{\texttt{\model} (\textbf{ours})} 
& \textbf{55.7} & \textbf{83.2} & \textbf{91.2} & \textbf{230.1} 
& \textbf{35.6} & \textbf{56.5} & \textbf{68.1} & {\textbf{160.2}} 
& \textbf{55.4} & \textbf{83.4} & \textbf{92.4} & {\textbf{231.2}} 
& \textbf{56.0} & \textbf{81.9} & \textbf{87.8} & {\textbf{225.7}} \\

\bottomrule
\end{tabular}
}
\label{table:4datasets}
\vspace{-0.5em}
\end{table*}

\begin{table*}[!t]
\footnotesize
\centering
\caption{\textbf{Out-of-distribution} performance of \textbf{T2V} models on {LSMDC}, {ActivityNet}, and {DiDeMo}. The best and second-best results are highlighted in \textbf{bold} and \underline{underlined}, respectively.}
\vspace{-0.7em}
\label{tab:ood}
 \setlength{\tabcolsep}{9pt}
\resizebox{1.0\linewidth}{!}{
\begin{tabular}{l|cccc|cccc|cccc}
\toprule
\multicolumn{1}{c|}{\multirow{2}{*}{\textbf{Methods}}} 
& \multicolumn{4}{c|}{\textbf{{MSR-VTT$\rightarrow$} {LSMDC}}} 
& \multicolumn{4}{c|}{\textbf{{MSR-VTT $\rightarrow$} \textbf{ActivityNet}}} 
& \multicolumn{4}{c}{\textbf{{MSR-VTT $\rightarrow$} {DiDeMo}}} \\ 

& {R@1$\uparrow$} & {R@5$\uparrow$} & {R@10$\uparrow$} & \multicolumn{1}{c|}{Rsum$\uparrow$}  & {R@1$\uparrow$} & {R@5$\uparrow$} & {R@10$\uparrow$} & \multicolumn{1}{c|}{Rsum$\uparrow$}  & {R@1$\uparrow$} & {R@5$\uparrow$} & {R@10$\uparrow$} & {Rsum$\uparrow$} \\ 
\midrule

\texttt{CLIP4Clip} & 15.3 & 31.3 & 40.5 & 87.1 & 29.1 & 58.3 & 72.1 & 159.5 & 31.8 & 57.0 & 66.1 & 154.9 \\

\texttt{EMCL-Net} & 16.6 & 29.3 & 36.5 & 82.4 & 28.7 & 56.8 & 70.6 & 156.1 & 30.0 & 56.1 & 65.8 & 151.9 \\

\texttt{DiffusionRet} & {17.1} & {32.4} & {41.0} & {90.5} & {31.5} & {60.0} & {73.8} & {165.3} & {33.2} & {59.3} & {68.4} & {160.9}  \\

\rowcolor{beaublue!50} \textbf{\texttt{\model} (\textbf{ours})} & \textbf{24.0} & \textbf{45.0} & \textbf{56.8} & \textbf{125.8}  & \textbf{35.2} & \textbf{64.8} & \textbf{89.0} & \textbf{189.0} & \textbf{36.7} & \textbf{62.6} & \textbf{71.2} & \textbf{170.5} \\

\bottomrule
\end{tabular}
}
\vspace{-1.5em}
\end{table*}

\begin{table}[t]
\caption{{T2V} {ablation studies} for network designs on {MSR-VTT}. Exp \#1 is our baseline which is \texttt{CLIP4clip} with {WTI} matching.}
\label{table:overall_ablation}
\vspace{-0.7em}
\setlength{\tabcolsep}{2pt}
\resizebox{\linewidth}{!}{%
\begin{tabular}{l|c|c|c|c|ccc}
\toprule[1.25pt]
 \textbf{Exp} & \multicolumn{3}{c|}{\textbf{Visual Scene Debias}}             & \textbf{Textual Content Debias}  & \multicolumn{3}{c}{\textbf{Performance}}  \\ 
 \hline
 & \multicolumn{2}{c|}{Scene Element} & \multicolumn{1}{c|}{\multirow{3}{*}{\shortstack{Captioning \\Head}}} & \multirow{3}{*}{\shortstack{Content-Bias\\ Representation\\ Disentanglement}} & \multirow{3}{*}{R@1$\uparrow$} & \multirow{3}{*}{R@5$\uparrow$} & \multirow{3}{*}{R@10$\uparrow$} \\ \cline{2-3}
 & \shortstack{Scene\\Entities} & \shortstack{Scene\\Activities} &  &  &  & &   \\ \midrule 
 

 \#1 & \xmark & \xmark & \xmark & \xmark &  45.7 &  73.0 &  82.6 \\ 

 \#2 & \cmark & \xmark & \xmark & \xmark &  49.2 & 75.3 & 83.0 \\ 
 \#3 & \xmark & \cmark & \xmark & \xmark &  49.4 & 75.7 & 83.4 \\ 
 \#4 & \cmark & \cmark & \xmark & \xmark &  51.5 & 76.6 & 84.8 \\ 
 \#5 & \cmark & \cmark & \cmark & \xmark &  52.1 & 77.2 & 86.0 \\ 
\midrule
 \#6 & \xmark & \xmark & \xmark & \cmark & 47.0 & 75.8 & 84.0 \\ 
\midrule
\rowcolor{beaublue!50} \#7 & \cmark & \cmark & \cmark & \cmark & \textbf{53.5} & \textbf{78.6} & \textbf{86.5}  \\ 
 \bottomrule[1.25pt]
\end{tabular}}
\end{table}


\subsection{Experimental Setups}

\noindent\textbf{Datasets.} We conducted experiments on five major TVR benchmarking datasets:
\begin{itemize}
\item (1) MSR-VTT~\cite{DBLP:conf/cvpr/XuMYR16} dataset contains 10,000 YouTube videos, with 20 descriptions per video. We follow the 1k-A split~\cite{DBLP:journals/ijon/LuoJZCLDL22} to conduct training on 9,000 videos and report testing results on the other 1,000 videos. 

\item (2) MSVD~\cite{DBLP:conf/acl/ChenD11} contains 1,970 videos with over 80,000 descriptions, with 40 descriptions on average per video. We follow the official split with 1,200 videos for training and 670 videos for testing. 

\item (3) LSMDC~\cite{lsmdc2016MovieAnnotationRetrieval} contains 118,081 video clips, which are extracted from 202
movies. We follow the split of \cite{DBLP:conf/eccv/Gabeur0AS20} with 100,000 videos for training and 1,000 videos for testing. 

\item (4) ActivityNet Captions~\cite{DBLP:conf/iccv/KrishnaHRFN17} contains 20,000 YouTube videos, following the training and evaluation protocol in~\cite{DBLP:journals/ijon/LuoJZCLDL22}, we report results on the “val1” split of 10,009 and 4,917 as the train and test set, respectively. 
\item (5) DiDeMo~\cite{DBLP:conf/iccv/HendricksWSSDR17} contains 10,464 videos with over 40,543 text descriptions. We follow the training and evaluation protocol in~\cite{DBLP:journals/ijon/LuoJZCLDL22}.
\end{itemize}
These datasets vary in video duration, research goals, target users, and text annotations, providing a comprehensive evaluation of different methods. 
We evaluate the {Text$\rightarrow$Video} (\textbf{T2V}) performance and provide additional results of {Video$\rightarrow$Text} (\textbf{V2T}) performance on standard rank-based
metrics \ie Recall at top \{1, 5, 10\} (recall at rank 1, 5, 10), Rsum (R@1 + R@5 + R@10). 

\vspace{0.5em}
\noindent\textbf{Implementation Details.} 
We use \texttt{CLIP} ViT/B-32 as the backbone for both \textit{Video Encoder} and \textit{Text Encoder}. 
We set $N_{t}=32$ and $N_{f}=32$ as the number of word tokens and video frames for all datasets except {DiDeMo} and {ActivityNet}, where $N_{t}$ and $N_{f}$ are set to 64. 
We train with batch size of 128 for 5 epochs, except for {DiDeMo} with 10 epochs and {ActivityNet} with 20 epochs. We use the Adam~\cite{DBLP:journals/corr/KingmaB14} as the optimizer. The learning rate follows the cosine schedule with a linear warmup strategy ~\cite{DBLP:journals/corr/GoyalDGNWKTJH17}. 
For Equation~\ref{eq:top-k}, we set $\kappa=20$. For Equation~\ref{eq:total_loss}, We set $\lambda_{\texttt{cap}} = 0.3$, $\lambda_{\texttt{rec}}=0.5$, and $\lambda_{\mathbb{KL}} = 1e^{-4}$.

\subsection{Quantitative Results}
\noindent\textbf{Comparison with SOTA.} 
We compare the proposed \model with SOTA methods. Table~\ref{table:msrvtt} shows the retrieval results on the {MSR-VTT} dataset, where our proposed method attains SOTA on both \textbf{T2V} and \textbf{V2T} tasks. We achieve 53.5 in the \textbf{T2V} task, outperforming the runner-up \texttt{TextProxy}~\cite{xiao2025textproxy} by 1.2 and the third best \texttt{DITS}~\cite{DBLP:conf/nips/WangWLGDRRT24} by 1.6 w.rt. R@1. On other metrics, \model also significantly outperforms the runner-up results. {In \textbf{V2T} task, we achieve 52.2, outperforming the runner-up \texttt{NarVid}~\cite{hur2025narratingthevideo} by 2.2 w.rt. R@1.}
Table~\ref{table:4datasets} shows the retrieval results on \textbf{T2V}, where \model consistently achieves the best results on all the metrics on all four datasets {MSVD}, {LSMDC}, {ActivityNet}, and {DiDeMo}. It demonstrates that our strategy can work well across different domains and different text-video data variations, thus underscoring the efficacy of \model through the ``Visual-Linguistic Bias Mitigation'' technique. For example, on {MSVD}, our approach at 55.7 has outperformed the recent SOTA method \texttt{NarVid}~\cite{hur2025narratingthevideo} by 2.6 w.r.t R@1. {Similarly on {ActivityNet}, \model significantly outperforms the runner-up \texttt{TextProxy} by 2.4 and w.r.t R@1, achieving SOTA performance of 55.4.}

\vspace{0.5em}
\noindent\textbf{Generalization to Unseen Domains.}
Our mitigation of visual-linguistic biases is geared towards maximizing the {quality} of cross-modal semantic representations during training, thereby capturing better underlying patterns that can generalize to unseen data. Hence, to evaluate {\model}'s debias capability, we evaluate \model on out-of-distribution retrieval settings~\cite{DBLP:conf/cvpr/ChenZJW20} (denoted as {A} $\rightarrow$ {B}) where the model is trained on dataset {A} and benchmarked on dataset {B}, which is unseen during training. In Table~\ref{tab:ood}, we compare our {\model} with recent SOTAs and the baseline \texttt{CLIP4Clip} in three OOD retrieval benchmarks ({MSR-VTT} $\rightarrow$ {LSMDC}, {MSR-VTT} $\rightarrow$ {ActivityNet}, and {MSR-VTT} $\rightarrow$ {DiDeMo}) where \model significantly outperforms others on the three benchmarks. These results suggest that our bias mitigation capability supports model generalization and transfer to unseen data.

\begin{figure}[ht]
\scriptsize
\captionsetup[subfigure]{labelformat=empty}
\centering
\begin{subfigure}{0.325\linewidth}
    \begin{tikzpicture}
    \begin{axis}[
            xlabel = $\lambda_{\text{Cap}}$,
            ylabel = R@1,
            ymax = 54,
            ymin = 50,
            xtick = {0.1, 0.2, 0.3, 0.4, 0.5},
            xticklabels={0.1, 0.2, 0.3, 0.4, 0.5},
            width=1.3\linewidth,
            height=0.42*\axisdefaultheight,
            legend style={at={(2.0,1.45)},anchor=north,legend columns=-1}]
            \addplot[red,fill=none,mark=*, fill opacity=0.2] table[x=x, y=T2V] {charts/lambdacap.txt};
            \addlegendentry{Text$\rightarrow$Video (T2V)}
            \addplot[black,fill=none,mark=triangle*, fill opacity=0.2] table[x=x, y=V2T] {charts/lambdacap.txt};
            \addlegendentry{Video$\rightarrow$Text (V2T)}
        \end{axis}
    \end{tikzpicture}
    \caption{}
    \label{subfig:lambdacap}
\end{subfigure}
\begin{subfigure}{0.325\linewidth}
    \begin{tikzpicture}
        \begin{axis}[
            xlabel = $\lambda_{\text{Rec}}$,
            ylabel = R@1,
            ymax = 54,
            ymin = 50,
            xtick = {0.3, 0.4, 0.5, 0.6, 0.7},
            xticklabels={0.3, 0.4, 0.5, 0.6, 0.7},
            width=1.3\linewidth,
            height=0.42*\axisdefaultheight,
            legend style={at={(4.32,1.45)},anchor=north,legend columns=-1}]
            \addplot[red,fill=none,mark=*, fill opacity=0.2] table[x=x, y=T2V] {charts/lambdarec.txt};
            \addplot[black,fill=none,mark=triangle*, fill opacity=0.2] table[x=x, y=V2T] {charts/lambdarec.txt};
        \end{axis}
    \end{tikzpicture}
    \caption{}
    \label{subfig:lambdarec}
\end{subfigure}
\begin{subfigure}{0.325\linewidth}
    \begin{tikzpicture}
        \begin{axis}[
            xlabel = $\lambda_{\mathbb{KL}} (\cdot 10^{-4})$,
            ylabel = R@1,
            ymax = 54,
            ymin = 50,
            xtick = {1, 2, 3, 4, 5},
            xticklabels={1, 2, 3, 4, 5},
            width=1.3\linewidth,
            height=0.42*\axisdefaultheight,
            legend style={at={(4.32,1.45)},anchor=north,legend columns=-1}]
            \addplot[red,fill=none,mark=*, fill opacity=0.2] table[x=x, y=T2V] {charts/lambdaKL.txt};
            \addplot[black,fill=none,mark=triangle*, fill opacity=0.2] table[x=x, y=V2T] {charts/lambdaKL.txt};
        \end{axis}
    \end{tikzpicture}
    \caption{}
    \label{subfig:lambdaKL}
\end{subfigure}
\vspace{-3.5em}
\caption{{Hyperparameter sensitivity }study on {MSR-VTT}.}
\label{fig:hyperparameters}
\vspace{-1.0em}
\end{figure}
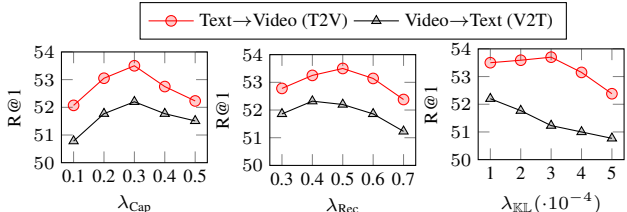

\vspace{0.5em}
\noindent\textbf{Hyperparameter Sensitivity.} 
In Figure~\ref{fig:hyperparameters}, we evaluate hyperparameters $\lambda_{\mathrm{Cap}} \in [0.1, 0.5]$, $\lambda_{\mathrm{Rec}} \in [0.3, 0.7]$ and $\lambda_{\mathbb{KL}} \in [0.0001, 0.0005]$. 
In Figure~\ref{fig:hyperparameters}(left), \model achieves highest R@1 score when $\lambda_{\mathrm{Cap}} = 0.3$ for \textbf{T2V}. As a result, we select $\lambda_{\mathrm{Cap}} = 0.3$. 
In Figure~\ref{fig:hyperparameters}(middle), the best is achieved with $\lambda_{\mathrm{Rec}}=0.5$, so we set $\lambda_{\mathrm{Rec}}=0.5$ as the default.
In Figure~\ref{fig:hyperparameters}(right), we show that $\lambda_{\mathbb{KL}}$ is highly sensitive and leads to a trade-off between \textbf{T2V} and \textbf{V2T} performance when $\lambda_{\mathbb{KL}}$ change from 0.0001 to 0.0003. Thus, we set $\lambda_{\mathbb{KL}}=0.0001$ to balance between \textbf{T2V} and \textbf{V2T}. 

\vspace{0.5em}
\noindent\textbf{Effect of the Visual Scene Debias.} 
Table~\ref{table:overall_ablation} shows that scene element features can improve performance through (i) \textit{Scene Entities} (see Exp \#2) or \textit{Scene Activities} (see Exp \#3):---either when only one type of scene element is fused with video representation, the R@1 score improved by a large margin of 1.4 (only scene entities) and 1.7 (only scene activities) compared to Baseline;---or, through \textit{Scene Elements Aggregation} (see Exp \#4), these features can be combined to enable an improvement over single feature utilization, with R@1 increased to 51.5 from 49.2 / 49.4. Furthermore, through (ii) \textit{Captioning Head} (see Exp \#5), better alignment between element-aware visual scene features and textual features can also be achieved, hinting that visual embeddings are enabled with finer-grained cross-modal features as R@1 increased from 51.5 to 52.1.

\vspace{0.5em}
\noindent\textbf{Effect of Textual Content Debias.}
Table~\ref{table:overall_ablation} shows the effect of the \textit{Textual Content Debias} module. 
From Exp \#5 and \#7, we observe that the efficacy of the \textit{Text Content Debias} and \textit{Visual Scene Debias} improves R@1 from 52.1 to 53.5.
This suggests that the \textit{Textual Content Debias} module is complementary with \textit{Visual Scene Debias} to improve performance. 

\vspace{0.5em}
\noindent\textbf{Effect of Textual Bias Features.}
To evaluate the impact of bias on textual representations in retrieval tasks, we conduct a controlled analysis by fusing content and bias features using a weighted sum: $\hat{\mathbf{t}}^{(i)} + \alpha \cdot \tilde{\mathbf{t}}^{(i)}$, where $\hat{\mathbf{t}}^{(i)}$ denotes the content embedding and $\tilde{\mathbf{t}}^{(i)}$ the bias embedding. The scalar parameter $\alpha$ modulates the degree of bias injected into the content representation. By varying $\alpha$, we systematically assess how retrieval performance degrades as bias intensity increases, thereby quantifying the trade-off between preserving semantic fidelity and suppressing bias-induced noise.

\begin{figure}[!htb]
    \centering
    \begin{minipage}{.47\linewidth}
       \centering
        \begin{tikzpicture}
        \begin{axis}[
            xlabel = $\alpha$,
            ylabel = R@1,
            ymax = 54,
            ymin = 41,
            xtick = {0.0, 0.2, 0.4, 0.6, 0.8, 1.0},
            xticklabels={0.0, 0.2, 0.4, 0.6, 0.8, 1.0},
            width=1.05\linewidth,
            height=0.45*\axisdefaultheight,
            legend style={at={(0.5,1.45)},anchor=north,legend columns=-1, font=\footnotesize}
        ]
            \addplot[red,fill=none,mark=*, fill opacity=0.2] table[x=x, y=T2V] {charts/bias_alpha.txt};
            \addlegendentry{(V2T)}
            \addplot[black,fill=none,mark=triangle*, fill opacity=0.2] table[x=x, y=V2T] {charts/bias_alpha.txt};
            \addlegendentry{(V2T)}
        \end{axis}
    \end{tikzpicture}
    \vspace{-1em}
    \caption{Effect of bias features on MSR-VTT. }
    \label{fig:bias_alpha}
    \end{minipage}%
    \hspace{5pt}
    \begin{minipage}{0.47\linewidth}
        \centering
        \includegraphics[width=1.\linewidth]{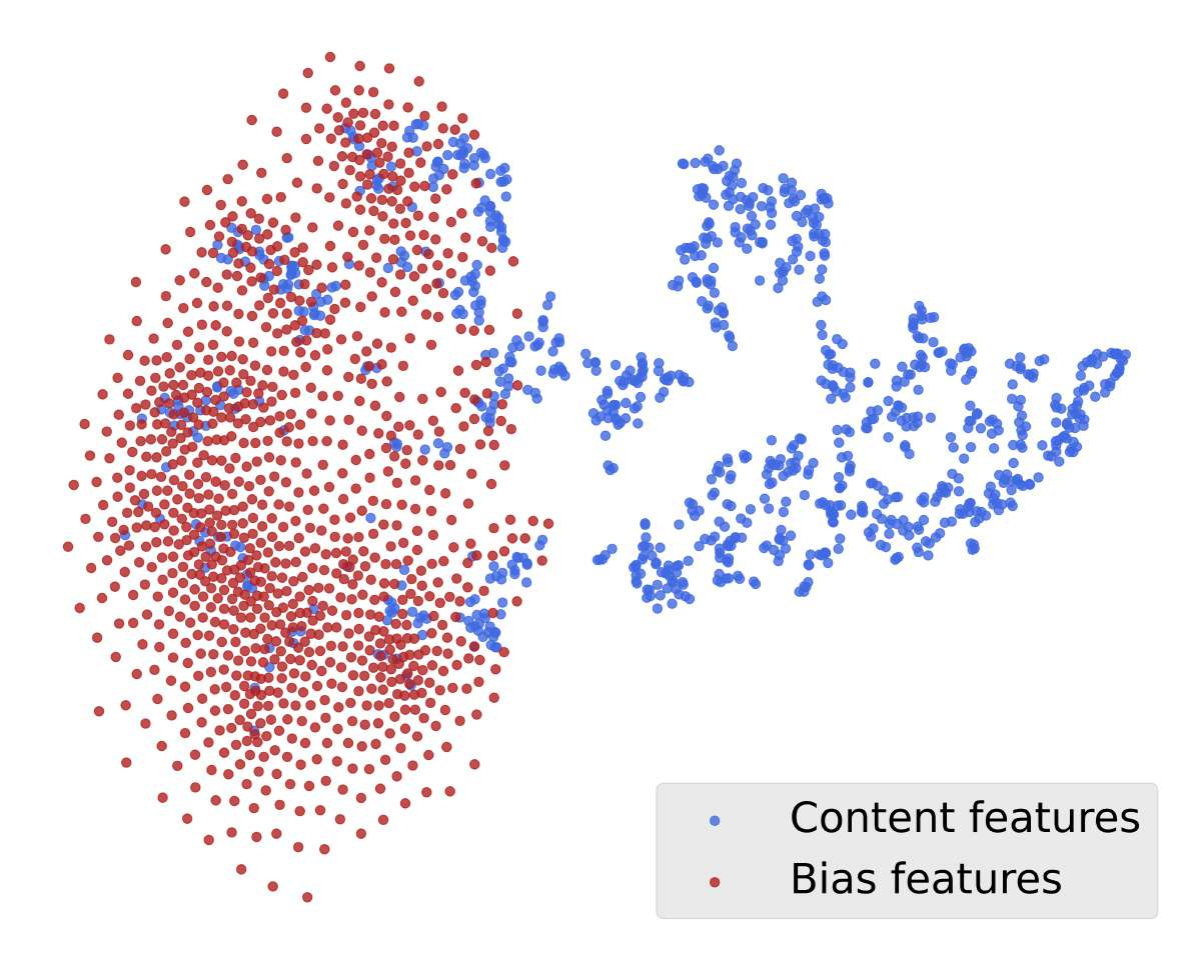}
    \caption{{t-SNE~\cite{vanDerMaaten2008} visualization between content and bias embeddings on MSR-VTT testset}.}
        \label{fig:content_bias_tsne}
    \end{minipage}
\end{figure}

Quantitatively, in Figure~\ref{fig:bias_alpha}, we illustrate the impact of varying \( \alpha \) on retrieval performance. As \( \alpha \) transitions from content-only to a mix of content and bias features, we observe that introducing more bias into the content increases noise during retrieval, which degrades performance. This demonstrates how the presence of bias disturbs the retrieval process.

As shown in Table~\ref{tab:content_bias_retrieval}, we further validate the importance of disentangling content features from bias features in the TVR task. Specifically, the use of content features $\hat{\mathbf{t}}^{(i)}$ achieves significantly higher retrieval performance compared to the use of bias features $\tilde{\mathbf{t}}^{(i)}$ across all recall metrics. When relying on $\hat{\mathbf{t}}^{(i)}$ (content features), the model attains an R@1 of 53.5\%, R@5 of 78.6\%, and R@10 of 86.5\%. In stark contrast, using $\tilde{\mathbf{t}}^{(i)}$ (bias features) alone results in extremely poor performance, with R@1 dropping to 1.3\%, R@5 to 10.2\%, and R@10 to 18.9\%. This substantial gap clearly illustrates that bias features, when isolated, fail to provide sufficient semantic grounding for accurate retrieval. Instead, they appear to introduce noise that misleads the model away from the true video content, corroborating our earlier hypothesis that biases embedded in the linguistic patterns can harm retrieval accuracy if not properly handled. Ultimately, our analysis underscores that effective bias mitigation, \eg using our proposed strategy, is essential for generalizable TVR.

Qualitatively, in Figure~\ref{fig:content_bias_tsne}, we visualize the content and bias feature embeddings of 1,000 samples from the MSR-VTT test set using t-SNE~\cite{vanDerMaaten2008}. The results show that our disentanglement process effectively separates content and bias components within the textual features. However, we also observe that some bias features remain partially entangled with the content representation. This highlights the inherent difficulty in completely disentangling bias information and suggests a promising direction for future investigation.

Additional ablation studies on \textbf{Effect of the Number of Scene Elements}, \textbf{Computational Cost}, and \textbf{Effect of Coefficient $g$} are included in the \textbf{Appendix}, respectively. 

\begin{table}[h]
    \centering
    \caption{Ablation study between content features $\hat{\mathbf{t}}$ or bias features $\tilde{\mathbf{t}}$ as textual features for T2V task on MSR-VTT.}
    \vspace{-0.7em}
    \label{tab:content_bias_retrieval}
 \setlength{\tabcolsep}{10pt}
    \resizebox{0.7\linewidth}{!}{
\begin{tabular}{l|ccc}
    \toprule
    \multicolumn{1}{c|}{\multirow{1}{*}{{Methods}}} & 
    \text{R@1$\uparrow$} & \text{R@5$\uparrow$} & \text{R@10$\uparrow$}\\ 
    \midrule
    \rowcolor{beaublue!50} \textit{using} $\hat{\mathbf{t}}$ (content) & \textbf{53.5} & \textbf{78.6} & \textbf{86.5}  \\
    
    \textit{using} $\tilde{\mathbf{t}}$ (bias) & 1.3 & 10.2 & 18.9\\ 
        \bottomrule
\end{tabular}
    }
\vspace{-1em}
\end{table}

\begin{figure*}[thb]
    \centering
    \includegraphics[width=1.\linewidth]{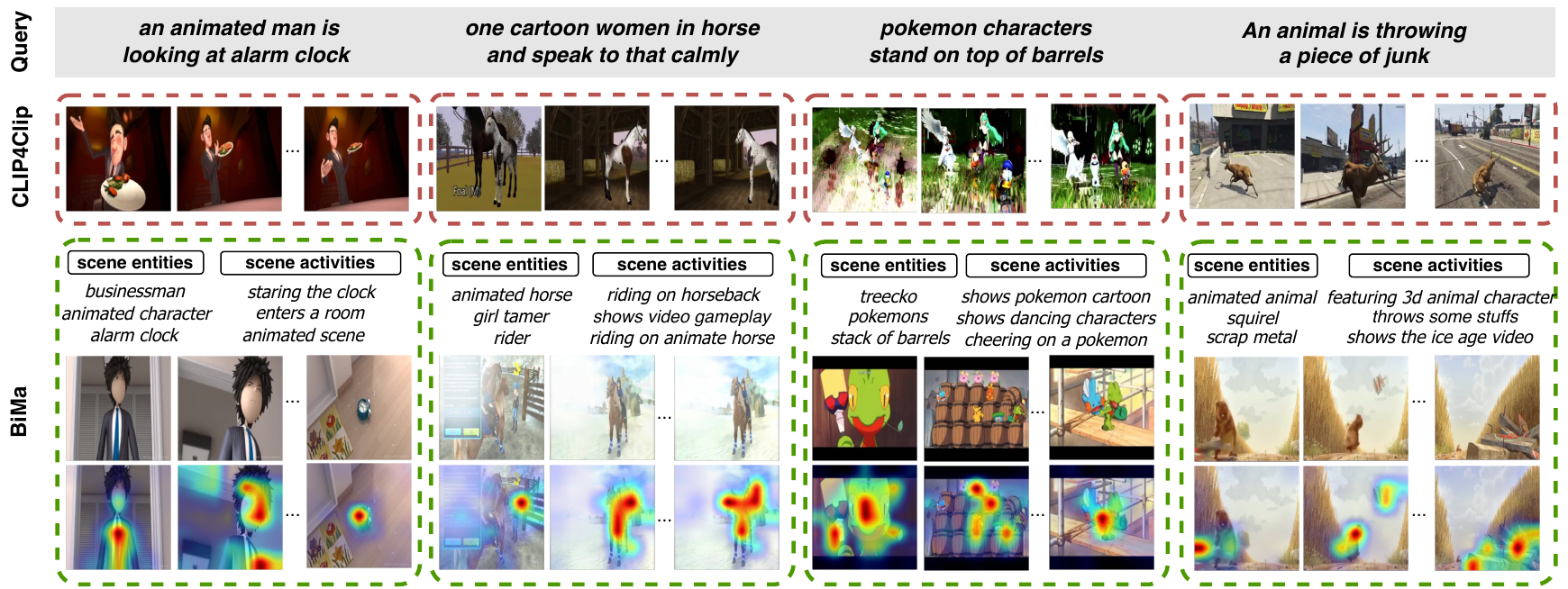} 
    \caption{\textbf{TVR qualitative Results.} Videos in green boxes are true top-1 videos retrieved from \model, and in red boxes are false top-1 videos retrieved from \texttt{CLIP4Clip}. We also provide attention visualizations and the top 3 relevant scene elements for each video to show where \model is focused on.}
    \label{fig:qualitative}
\end{figure*}


\subsection{Qualitative Results}
Figure~\ref{fig:qualitative} qualitatively demonstrates the performance of \model on MSR-VTT testset. We observe that \model is able to focus on relevant scene entities and scene activities, shown by the top three attention heatmaps per video, highlighting key elements in the scene. 
Further, \model can focus on small yet important actors and objects \eg, the ``alarm clock'' in the 1\textsuperscript{st} example and the ``animal'' in the 4\textsuperscript{th} example. Next, \model also can focus on behaviors \eg, ``throwing'' in the 4\textsuperscript{th} example.
This suggests that \model effectively integrates scene elements, which contributes to its improved retrieval accuracy over \texttt{CLIP4Clip}. In contrast, \texttt{CLIP4Clip} is unable to identify important scene elements and retrieves unrelated results. 
In summary, the experimental results highlight the advantage of \model in associating visual content with textual descriptions, showing its enhanced performance in capturing nuanced visual features relevant to the query.
Additional qualitative results are presented in the \textbf{Appendix}.

\section{Related Work}
\noindent\textbf{Text-Video Retrieval.}
TVR is a cross-modal retrieval task that aims to match videos with text descriptions~\cite{DBLP:conf/eccv/LiuXXCJ22,DBLP:conf/aaai/YanSGW0C023,jin2023video,Fang_2023_ICCV,jin2023diffusionret,DBLP:conf/aaai/ChenWLQMS23,teachclip}. Existing TVR methods typically leverage pre-trained VLMs like \texttt{CLIP}~\cite{DBLP:conf/icml/RadfordKHRGASAM21}, \texttt{BLIP}~\cite{DBLP:conf/icml/0001LXH22} and adopt contrastive learning on pairs of samples to enhance their cross-modal representations across vision and language domains. Recent advancements, such as \texttt{DiffusionRet}~\cite{jin2023diffusionret}, introduce multi-stage training mechanisms that use both discriminative and generative models to address challenges in out-of-domain retrieval tasks. In cases of ambiguous matching, where videos have multiple valid captions, uncertainty-based frameworks have emerged to handle this by learning a joint embedding space with probabilistic distance metrics. For example, \texttt{UATVR}~~\cite{Fang_2023_ICCV} aligns cross-modal features as a distribution-matching procedure, while \texttt{PAU}~\cite{DBLP:conf/nips/LiSGZS23} introduces semantic prototypes to capture ambiguous semantics within an uncertainty-based framework. Although our work also addresses data ambiguities, we distinguish between ``bias'' and ``uncertainty'' in terms of both definition and objective. We define ``bias'' as overlooked features stemming from coarse-grained data and annotator subjectivity, whereas ``uncertainty'' refers to the diversity within representations due to the variation of annotators. Our approach focuses on mitigating biases by recalibrating cross-modal features in the latent space and treating bias components as noise. In contrast, uncertainty-based methods attempt to model representation diversity by treating matched visual-linguistic representations as stochastic variables, and they do not directly alleviate visual-linguistic biases.

\noindent\textbf{Bias in Vision-Language Models.}
Bias in VLMs has become a growing concern as these models, widely used in real-world applications, are often pre-trained on large internet-scale datasets.
While this provides VLMs with extensive knowledge, it also makes them susceptible to inheriting biases present in the underlying data, such as cultural stereotypes, racial biases, and gender imbalances~\cite{bolukbasi2016man,garg2018word,liang2020towards,sun2019mitigating}. For instance, \texttt{CLIP} and \texttt{BLIP} tend to amplify societal biases~\cite{DBLP:conf/cvpr/WangQKGNHR20}. 
Recent literature explicitly categorizes these biases as representation biases—systematic deviations or skews within datasets, causing models to disproportionately rely on certain dataset-specific patterns rather than generalizable, task-relevant features~\cite{shvetsova2025unbiasing, liu2024decade}. These biases typically manifest as concept bias (where models rely heavily on prominent visual concepts), temporal bias (where temporal aspects are inadequately represented), and textual bias (arising from subjective annotator interpretations or emotionally charged annotations)~\cite{shvetsova2025unbiasing}. Additionally, recent empirical evaluations reveal that even modern datasets, designed with significant diversification efforts, still contain intrinsic dataset-specific biases, limiting models' generalizability and robustness~\cite{liu2024decade}.
As VLMs become more prominent, addressing and mitigating these biases is critical for ensuring ethical and fair AI applications. For visual bias, existing approaches have relied on costly solutions such as simulators~\cite{leclerc20223db} or crowdsourcing~\cite{idrissi2022imagenet,plumb2021finding} to annotate visual features. These approaches are not scalable and lack universal applicability. Recent studies~\cite{vo2023aoe,yamazaki2023vltint} utilize scene factorization to eliminate biases in visual representation by focusing on relevant scene entities. However, many of these approaches rely heavily on computationally intensive object detectors, restricting their applicability to unseen scenarios.
For textual bias, existing studies address biases in sentence embeddings by introducing training constraints~\cite{huang2019reducing} or directly modifying datasets~\cite{zhao2019gender}. Recent Autoencoder-based methods~\cite{DBLP:conf/eccv/ChenZYFWJL18,DBLP:conf/acl/JohnMBV19, DBLP:conf/cvpr/GuoLYLL19,DBLP:conf/mm/TanL0ZWCW22} aim to detach the domain bias from textual representations. However, these methods typically rely on supervised learning using additional bias annotations, limiting their generalizability and scalability.
To the best of our knowledge, \model represents a pioneering effort in systematically addressing these identified visual and textual representation biases in VLMs through the integration of scene element aggregation, visual attention redirection, and textual disentanglement. Crucially, our method is self-supervised and does not require additional human annotations, effectively addressing the limitations noted in prior studies~\cite{shvetsova2025unbiasing, liu2024decade}. 

\vspace{-0.5em}
\section{Conclusion}
\vspace{0.5em}
\noindent\textbf{Conclusion.} In this study, we addressed visual-linguistic bias challenges in the TVR task, drawing upon recent bias representation findings~\cite{liu2024decade,shvetsova2025unbiasing} to address an unexplored area in the TVR literature. We proposed \texttt{\model}, a novel framework to mitigate biases in both visual and textual representations.
Our proposed framework incorporates three modules-- \textit{Scene Element Construction}, \textit{Visual Scene Debias}, and \textit{Textual Content Debias}. Through extensive experiments and ablation studies across five major TVR benchmark datasets, we demonstrated the remarkable performance of our approach, surpassing existing SOTA TVR methods. Additionally, our \model exhibited remarkable performance in handling out-of-distribution retrieval problems showing the bias mitigation and generalization capabilities on unseen data variations.

\vspace{0.5em}
\noindent\textbf{Limitation.} Although our taxonomy dictionary is constructed from diverse sources containing a comprehensive set of scene elements, some inherent biases may remain unmitigated. Addressing these challenges remains an avenue for future research.

\vspace{0.5em}
\noindent\textbf{Broader Impacts.}
BiMa advances generalize and robust TVR by mitigating dataset-specific biases through innovative, self-supervised techniques. By ensuring models focus on meaningful content rather than superficial features, this work not only boosts retrieval accuracy but also sets a new standard for developing generalized multimedia systems--paving the way for more flexible AI applications in diverse real-world scenarios. 

\clearpage
\newpage
{
    \footnotesize
    \bibliographystyle{ieeenat_fullname}
    \bibliography{main}
}
\clearpage
\newpage
\setcounter{table}{0}
\renewcommand\thetable{\Alph{section}.\arabic{table}}
\setcounter{figure}{0}
\renewcommand\thefigure{\Alph{section}.\arabic{figure}}

\appendix

\begin{center}
{\Large \textbf{Appendix}}
\end{center}

\begin{table}[h]
\centering
\caption{Effect of the number of scene elements on MSR-VTT.}
\label{table:vocab-quant}
\setlength{\tabcolsep}{4pt}
\begin{tabular}{@{}c|ccc|ccc@{}}
\toprule
\multirow{2}{*}{\textbf{$\kappa$}} & \multicolumn{3}{c|}{ \textbf{Scene Entities}} & \multicolumn{3}{c}{\textbf{Scene Activities}} \\ 

& \multicolumn{1}{c}{R@1$\uparrow$} & \multicolumn{1}{c}{R@5$\uparrow$} & \multicolumn{1}{c|}{R@10$\uparrow$} & \multicolumn{1}{c}{R@1$\uparrow$} & \multicolumn{1}{c}{R@5$\uparrow$} & \multicolumn{1}{c}{R@10$\uparrow$} \\ 
\midrule

\textbf{$\kappa=10$} & 48.6 & 74.9 & 82.3 & 48.7 & 75.0 & 83.1  \\

\textbf{$\kappa=15$} & 48.9 & 74.9 & 82.6 & 48.9 & 75.2 & 83.1 \\

\rowcolor{beaublue!50} \textbf{$\kappa=20$} & \textbf{49.2} & \textbf{75.3} & \textbf{83.0} & \textbf{49.4} & \textbf{75.7} & \textbf{83.4}  \\

\textbf{$\kappa=25$} & 49.1 & 75.3 & 83.0 & 49.3 & 75.7 & 83.3 \\

\textbf{$\kappa=30$} & 48.9 & 75.2 & 82.8 & 48.9 & 75.6 & 83.4 \\

\bottomrule
\end{tabular}
\end{table}


\begin{table}[!t]
    \centering
    \caption{\textbf{Computational costs} of \model w.r.t. different $\kappa$ on MSR-VTT with a gallery size of 1,000 candidate videos and 1,000 text queries. The best and second best are \textbf{bold} and \underline{underlined}.}
    \label{table:computational_cost}
    \setlength{\tabcolsep}{4pt}
    \resizebox{1.\linewidth}{!}{
    \begin{tabular}{l|cc|c}
        \toprule
        \multicolumn{1}{c|}{\multirow{2}{*}{\textbf{Methods}}}
        & \multicolumn{2}{c|}{\textbf{Computational Cost}} 
        & \multicolumn{1}{c}{\textbf{Performance}} \\ 
        & \text{\shortstack{Inference Time (s)$\downarrow$}} & 
        \text{\shortstack{Memory (GB)$\downarrow$}}
        & 
        \text{\shortstack{R@1$\uparrow$}}  \\ 
        
        \midrule
        
        {\texttt{CLIP4Clip} (baseline)} & \textbf{31.12} & \textbf{4.1} & 45.7 \\
        \texttt{\model} ($\kappa=10$) & 36.25 & \underline{4.3} & 52.9 \\
        \texttt{\model} ($\kappa=15$) & 36.33 & \underline{4.3} & 53.2 \\
        \rowcolor{beaublue!50} \texttt{\model} ($\kappa=20$) & \underline{36.40} & \underline{4.3} & \textbf{53.5}  \\
        \texttt{\model} ($\kappa=25$) & 36.49 & \underline{4.3} & 53.3  \\
        \texttt{\model} ($\kappa=30$) & 36.52 & \underline{4.3} & 53.1 \\
        \bottomrule
    \end{tabular}}
\end{table}

\begin{table}[ht]
\centering
\caption{Ablation study on dataset combination of taxonomy dictionary.}
\label{table:num_dataset_taxonomy}
\setlength{\tabcolsep}{4pt}
\resizebox{1.0\linewidth}{!}{
\begin{tabular}{ccccc|ccc}
\toprule
\multicolumn{5}{c|}{\textbf{Dataset Source}} & \multicolumn{3}{c}{\textbf{T2V}} \\ 

\multicolumn{1}{c}{MSR-VTT} & \multicolumn{1}{c}{MSVD} & \multicolumn{1}{c}{LSMDC} & \multicolumn{1}{c}{ActivityNet} & \multicolumn{1}{c|}{DiDeMo} & \multicolumn{1}{c}{R@1$\uparrow$} & \multicolumn{1}{c}{R@5$\uparrow$} & \multicolumn{1}{c}{R@10$\uparrow$} \\ 
\midrule
\cmark & \xmark & \xmark & \xmark & \xmark & 49.7 & 75.7 & 83.8 \\
\cmark & \cmark & \xmark & \xmark & \xmark & 49.9 & 75.9 & 84.2 \\
\cmark & \cmark & \cmark & \xmark & \xmark & 50.4 & 76.1 & 84.6 \\
\cmark & \cmark & \cmark & \cmark & \xmark & 51.0 & 76.4 & 84.6 \\
\rowcolor{beaublue!50} \cmark & \cmark & \cmark & \cmark & \cmark & \textbf{51.5} & \textbf{76.6} & \textbf{84.8}\\

\bottomrule
\end{tabular}}
\end{table}

\begin{table}[ht]
    \centering
    \caption{Effectiveness of coefficient $g$ on Scene Elements Aggregation module on MSR-VTT.}
    \label{table:coefficient_g}
    \setlength{\tabcolsep}{8pt}
    \begin{tabular}{l|ccc}
        \toprule
        {\textbf{Method}} & 
        \text{R@1$\uparrow$} & \text{R@5$\uparrow$} & \text{R@10$\uparrow$} \\ 
        \midrule
        \textit{without} $g$ & 49.9 & 76.0 & 84.1 \\
        
        \rowcolor{beaublue!50} \textit{with} $g$ & \textbf{51.5} & \textbf{76.6} & \textbf{84.8} \\ 
        \bottomrule
    \end{tabular}
\end{table}




\section{Taxonomy Dictionary Details} 
\label{app:dictionary}
To construct a comprehensive taxonomy dictionary that serves as a thesaurus of scene elements—encompassing both entities and activities expressed as noun and verb phrases—we adopt a dual-stage strategy involving large-scale language and multimodal models. Specifically, we employ the Large Language Model (LLM) \textsc{Qwen2.5-7B}~\cite{qwen2.5,qwen2} to systematically extract phrasal expressions from video-text datasets, and the Large Multimodal Model (LMM) \textsc{LLaVA-OV-7B}~\cite{li2024llava} to enrich these expressions with visually plausible actions inferred from video content.

For phrasal extraction, we query the LLM \textsc{Qwen2.5-7B} using all captions available in existing TVR datasets with the prompt illustrated below. This prompt encourages the model to identify and group all possible \textit{noun phrases} and \textit{verb phrases} separately, while accounting for semantically equivalent variants:

\begin{tcolorbox}[boxrule=0pt, colframe=white, sharp corners, left=1mm, right=1mm, top=0.2mm, bottom=0.2mm]
\footnotesize
\textbf{PROMPT}
\\
List all possible noun phrases and verb phrases in the given caption. 
Present the two lists separately, separated by a semicolon (';'). Include alternative expressions for the same entity or action. For example, if the entity or action can be described in multiple ways, list them all. Ensure your output follows the format of the examples provided below.

\textbf{Caption:} In the heart of the kitchen, a man skillfully slices into a ripe mango, its golden flesh gleaming under the light.\\
\textbf{Verb phrases:} slicing a ripe mango; cutting into a ripe mango; gleaming under the light; shining in the light.\\
\textbf{Noun phrases:} the heart of the kitchen; a man; the man; a ripe mango; golden flesh; the light; the kitchen.
\\
\\
\textbf{Caption:} A woman sits by the fireplace, knitting a scarf as the flames crackle warmly in the background.\\
\textbf{Verb phrases:} sitting by the fireplace; sitting near the fireplace; knitting a scarf; knitting a piece of cloth; crackling in the background; burning in the background.\\
\textbf{Noun phrases:} a woman; the woman; the fireplace; a scarf; the flames; the fire; the background.
\end{tcolorbox}

To complement this text-based extraction, we employ the LMM \textsc{LLaVA-OV-7B} to analyze the raw visual content of videos in the same datasets. The model is prompted to generate a list of plausible short narrations in the form of verb-noun phrases (\eg, “slicing mango”), effectively capturing fine-grained, visually grounded actions that may not be explicitly mentioned in captions:

\begin{tcolorbox}[boxrule=0pt, colframe=white, sharp corners, left=1mm, right=1mm, top=0.2mm, bottom=0.2mm]
\footnotesize
\textbf{PROMPT}
\\
List all possible actions that could take place in the scene of the given video. Write each action as a short narration (a verb with a noun), separated by a semicolon (';'). Follow the structure shown in the examples below.

\textbf{Video:} \{video \#\} \\
\textbf{Narration:} Slicing mango; Holding knife; Cutting mango; Placing seed; Wiping counter; Dropping pieces; Gripping mango; Resting knife; Smelling mango; Gathering chunks.
\\
\\
\textbf{Video:} \{video \#\} \\
\textbf{Narration:} Knitting scarf; Holding needles; Looping yarn; Adjusting thread; Pulling stitch; Resting hands; Dropping yarn; Smelling smoke; Listening to flames; Rubbing hands; Folding scarf; Gathering wool; Staring at fire; Sitting still; Tapping needle.
\end{tcolorbox}

\begin{figure*}[t]
\centering
\begin{subfigure}{0.33\linewidth}
\centering
\includegraphics[width=1\linewidth]{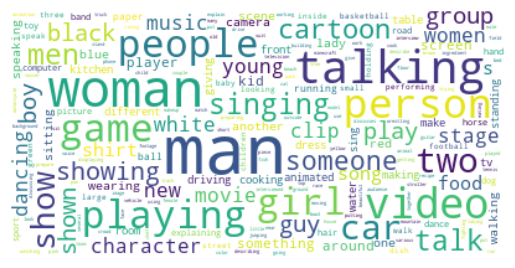}
\caption{\small{MSR-VTT} }
\end{subfigure}
\begin{subfigure}{0.33\linewidth}
\centering
\includegraphics[width=1\linewidth]{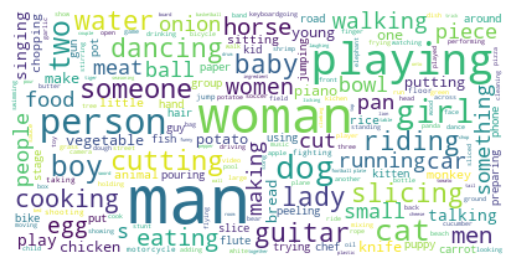}
\caption{\small{MSVD} }
\end{subfigure}
\begin{subfigure}{0.33\linewidth}
\centering
\includegraphics[width=1\linewidth]{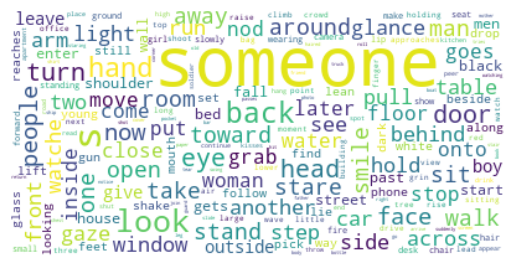}
\caption{\small{LSMDC} }
\end{subfigure}
\begin{subfigure}{0.33\linewidth}
\centering
\includegraphics[width=1\linewidth]{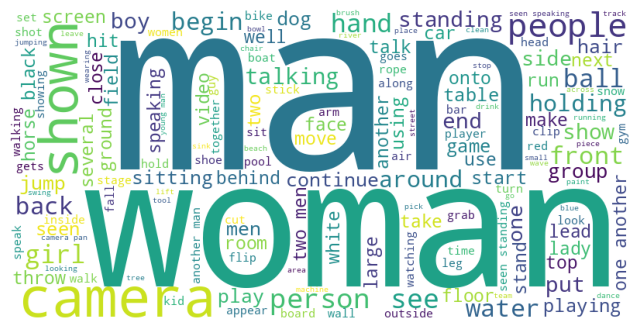}
\caption{\small{ActivityNet} }
\end{subfigure}
\begin{subfigure}{0.33\linewidth}
\centering
\includegraphics[width=1\linewidth]{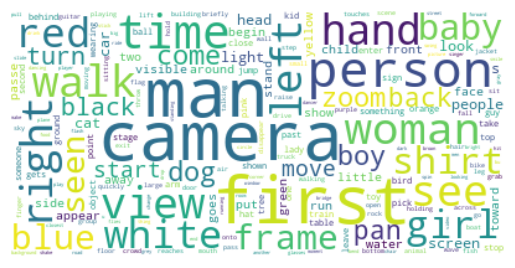}
\caption{\small{DiDeMo} }
\end{subfigure}
\caption{Word clouds of textual descriptions in different datasets.} 
\label{fig:word_cloud}
\end{figure*}

\begin{figure*}[!ht]
    \centering
    \includegraphics[width=1.\linewidth]{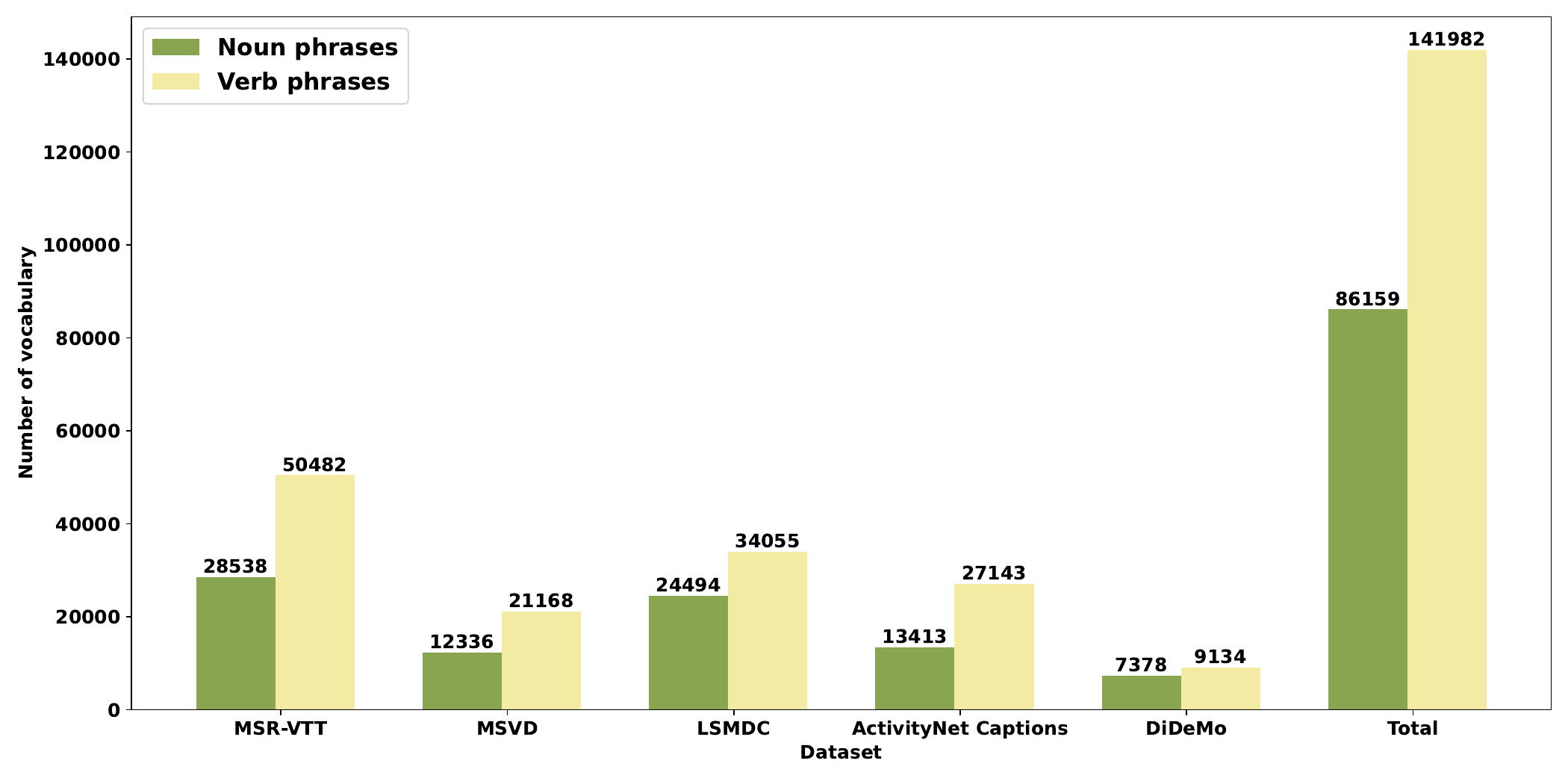} 
    \vspace{-0.5cm}\caption{\textbf{Taxonomy Dictionary.}}
    \label{fig:scene_elements_statistics}
\end{figure*}

After collecting these phrasal outputs, we apply deduplication to retain only unique noun and verb phrases. This process results in a diverse and comprehensive lexicon of scene elements, which we visualize as word clouds in Fig.~\ref{fig:word_cloud}, with each cloud representing a different dataset. As shown in Fig.~\ref{fig:scene_elements_statistics}, our method successfully extracts a total of 86,159 distinct noun phrases and 141,981 distinct verb phrases across five large-scale TVR benchmarks: {MSR-VTT}, {MSVD}, {LSMDC}, {ActivityNet Captions}, and {DiDeMo}.

For each dataset, we also provide examples of it in Table~\ref{tab:text_styles} to demonstrate the different variations of textual annotations, hinting at the textual biases that can be encountered.

\section{Scene Elements Details}
\label{app:scene_element}
The abundance of noun and verb phrases in video-text annotations introduces potential sources of inherent bias, as different textual descriptions for the same visual content can vary significantly in focus, vocabulary, and granularity. Consequently, models trained on such data may inadvertently overfit to specific linguistic patterns, leading to sub-optimal generalization. To address this, we construct a large-scale vocabulary of scene elements—spanning entities and actions—from a diverse collection of datasets to dilute such biases during training.

Our core hypothesis is that smaller or domain-specific dictionaries are more prone to encoding biases, whereas larger and more diverse vocabularies help reduce representational skew by incorporating a broader spectrum of semantic features. Empirical results from our ablation study (Table 4 - Main paper) validate this hypothesis: models utilizing Scene Elements consistently outperform those without, highlighting their effectiveness in mitigating bias.

To build this robust vocabulary, we aggregate scene elements from five large-scale TVR benchmarks: {MSR-VTT}, {MSVD}, {LSMDC}, {ActivityNet Captions}, and {DiDeMo}. As shown in Table~\ref{table:num_dataset_taxonomy}, increasing the number of datasets used to construct the dictionary yields consistent performance gains, further underscoring the importance of diverse and fine-grained video descriptions for reducing bias.

To further improve the quality and robustness of the taxonomy, we advocate for extending it to include additional datasets from varied domains. Furthermore, we propose exploring an open-vocabulary setting in which the dictionary is dynamically expanded during training to accommodate emerging or domain-specific scene elements. We leave the design of such adaptive and scalable vocabularies as a promising direction for future research.

Further, we observe that the number of verb phrases is approximately two times more than the number of noun phrases. This is because an actor can perform various activities, and annotators use different tenses and synonyms to describe activity(s), actor(s) in a video can engage in a list of behaviors.
Moreover, given that the number of verb phrases is more than the number of noun phrases, the verb phrases may inject more bias into TVR frameworks than noun phrases. Consequently, if we can mitigate biases that come from verb phrases, we may achieve better accuracy than mitigating biases that come from noun phrases. This aligns with the ablation study result (see Table~\ref{table:vocab-quant}). We observe that when solely integrating scene activities (i.e., verb phrases), we attain superior accuracy (with an increase of over 2\% on R@1 metrics) compared to when only integrating scene entities (i.e., noun phrases).

\section{Additional Ablation Studies}
\label{app:add_quant}
To further illustrate the capabilities of our \texttt{BiMa}, we conduct additional  quantitative analyses as follows:

\begin{itemize}
    \item Table~\ref{table:vocab-quant} illustrates the effectiveness of the number of scene elements used.
    \item Table~\ref{table:computational_cost} shows the efficiency of our method versus the baseline, across different number of scene elements.
    \item Table~\ref{table:coefficient_g} illustrates the effectiveness of the coefficient $g$. 
\end{itemize}

\noindent\textbf{Effect of the Number of Scene Elements.}
The number of scene elements $\kappa$ controls how many scene elements are fused into the video representation. In Table~\ref{table:vocab-quant}, the overall performance improves and then decreases. We observe that a small number of scene elements may limit the effect of the Visual Scene Debias module to focus on fine-grained information in the visual scene. However, a larger number of scene elements reduces the discriminability of the video representation. We achieve the best performance with $\kappa = 20$.

\noindent\textbf{Computational Cost.}
In practical scenarios, the taxonomy dictionary's integration is computationally efficient, as it matches each video with scene elements to find the top-$\kappa$ relevant scene elements only once. Table~\ref{table:computational_cost} presents comparison on computational cost and performance, detailing both inference time and memory usage. We observe that \model achieves higher accuracy than \texttt{CLIP4Clip} while only incurring a slight increase in inference time and maintaining comparable memory consumption. Additionally, as $\kappa$ increases, the impact on \model's computational cost becomes minimal. This experiment demonstrates that \model is computationally efficient and balances accuracy with resource usage effectively. For enhanced efficiency, the final fusion feature vectors between each video and top-$\kappa$ relevant scene elements can be directly indexed into the target database to reduce the fusion step during inference. 

\noindent\textbf{Effect of the coefficient $g$.} As depicted in Table~\ref{table:coefficient_g}, we show that coefficient $g$ can amplify the discriminative characteristic of each original feature by facilitating a seamless embedding aggregation process by acting as the association between each feature token and its most closely relevant token from the other feature tokens. After integrate the coefficient $g$ into the \textit{Scene Element Aggregation} module (in the main paper), the performance increase by 0.5\% w.r.t R@1.


\section{Additional Qualitative Visualizations}
\label{app:qualitative}
\begin{figure*}[!b]
\centering    \includegraphics[width=0.95\linewidth]{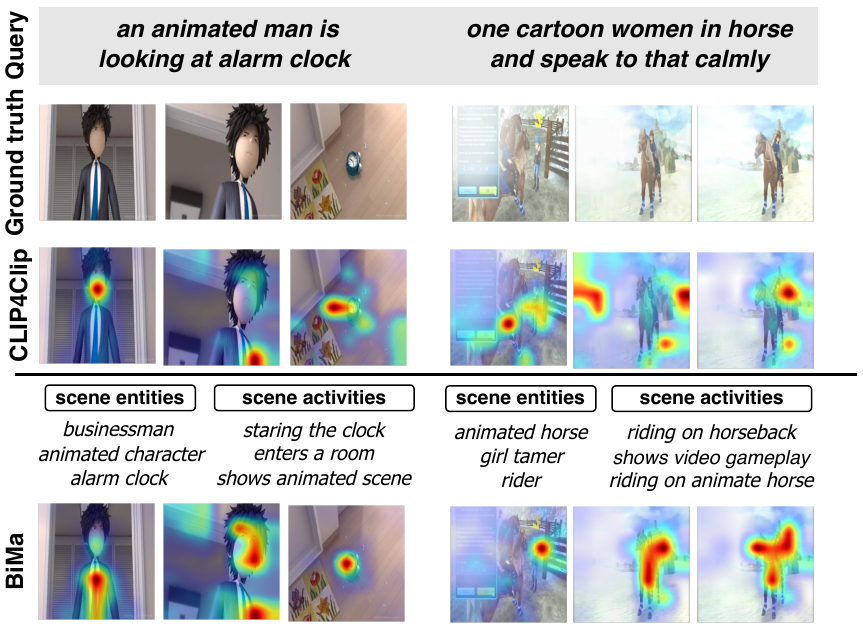} 
\vspace{-0.5cm}
    \caption{\textbf{Attention Map Results.} From top-to-bottom: Textual query description, ground truth video, \texttt{CLIP4Clip}'s attention map and  \texttt{BiMa}'s attention map.}
    \label{fig:attn_1}
\end{figure*}

\begin{figure*}[!ht]
\centering    \includegraphics[width=0.85\linewidth]{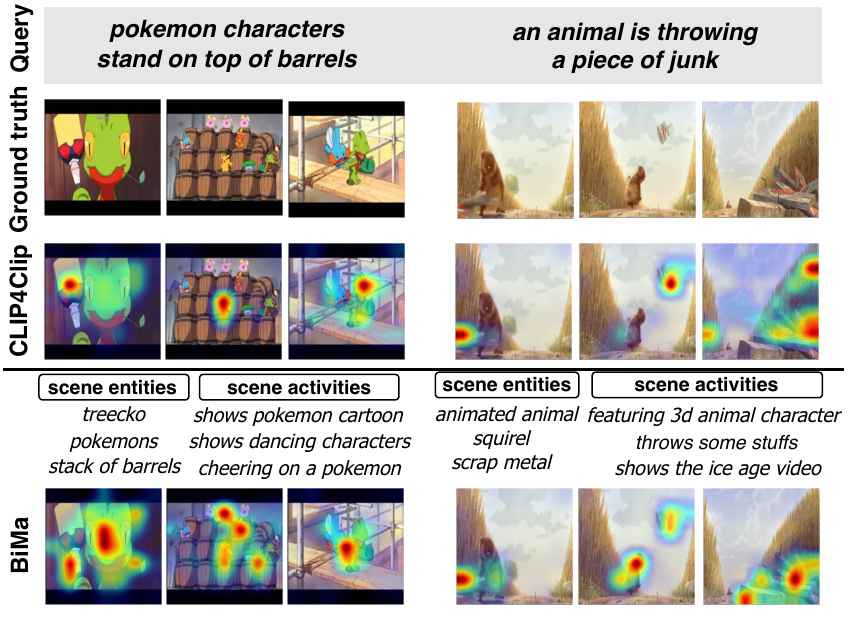} 
    \caption{\textbf{Attention Map Results.} From top-to-bottom: Textual query description, ground truth video, \texttt{CLIP4Clip}'s attention map and  \texttt{BiMa}'s attention map.}
    \label{fig:attn_2}
\end{figure*}

\begin{figure*}[!ht]
\centering    \includegraphics[width=0.95\linewidth]{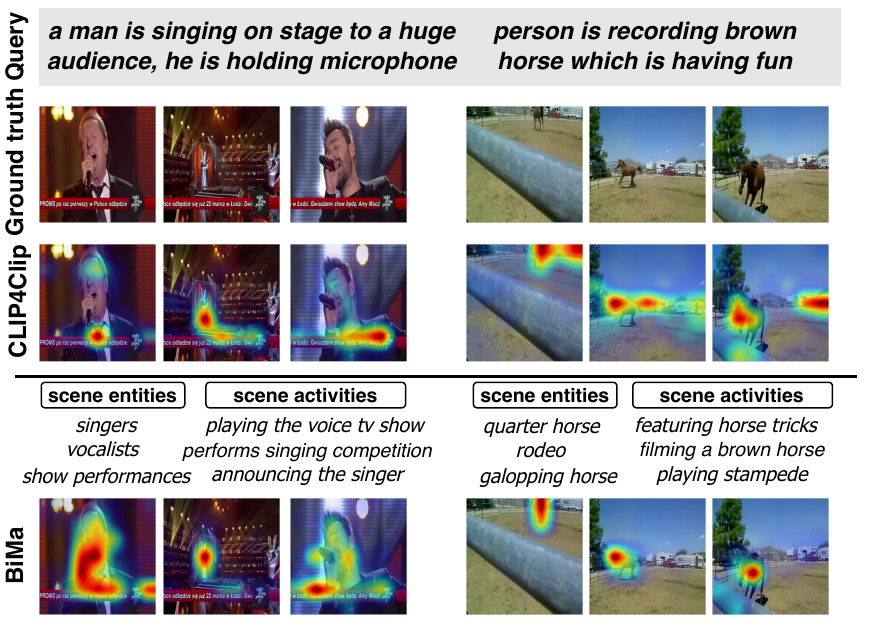} 
    \caption{\textbf{Attention Map Results.} From top-to-bottom: Textual query description, ground truth video, \texttt{CLIP4Clip}'s attention map and  \texttt{BiMa}'s attention map.}
    \label{fig:attn_3}
\end{figure*}

\begin{figure*}[!t]
    \centering
    \includegraphics[width=0.65\linewidth]{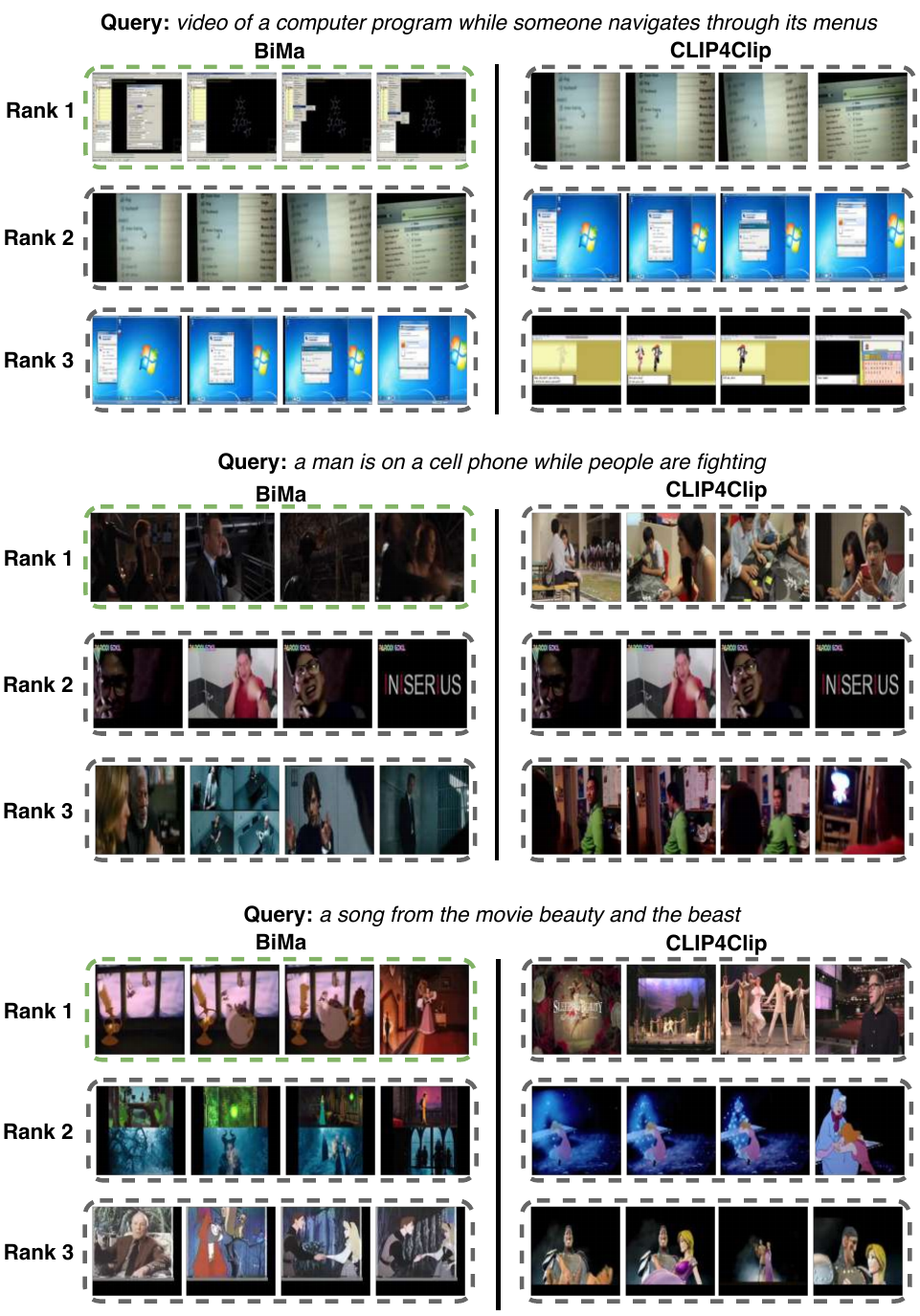} 
    \caption{\textbf{Text-to-video Results.} On the left are \texttt{BiMa}'s retrieval results and on the right are \texttt{CLIP4Clip}'s retrieval results. Ground truth videos are highlighted in the green box. These results demonstrate that our method can align the correlation between text and video effectively.}
    \label{fig:retrieval_1}
\end{figure*}

\begin{figure*}[!t]
    \centering
    \includegraphics[width=0.65\linewidth]{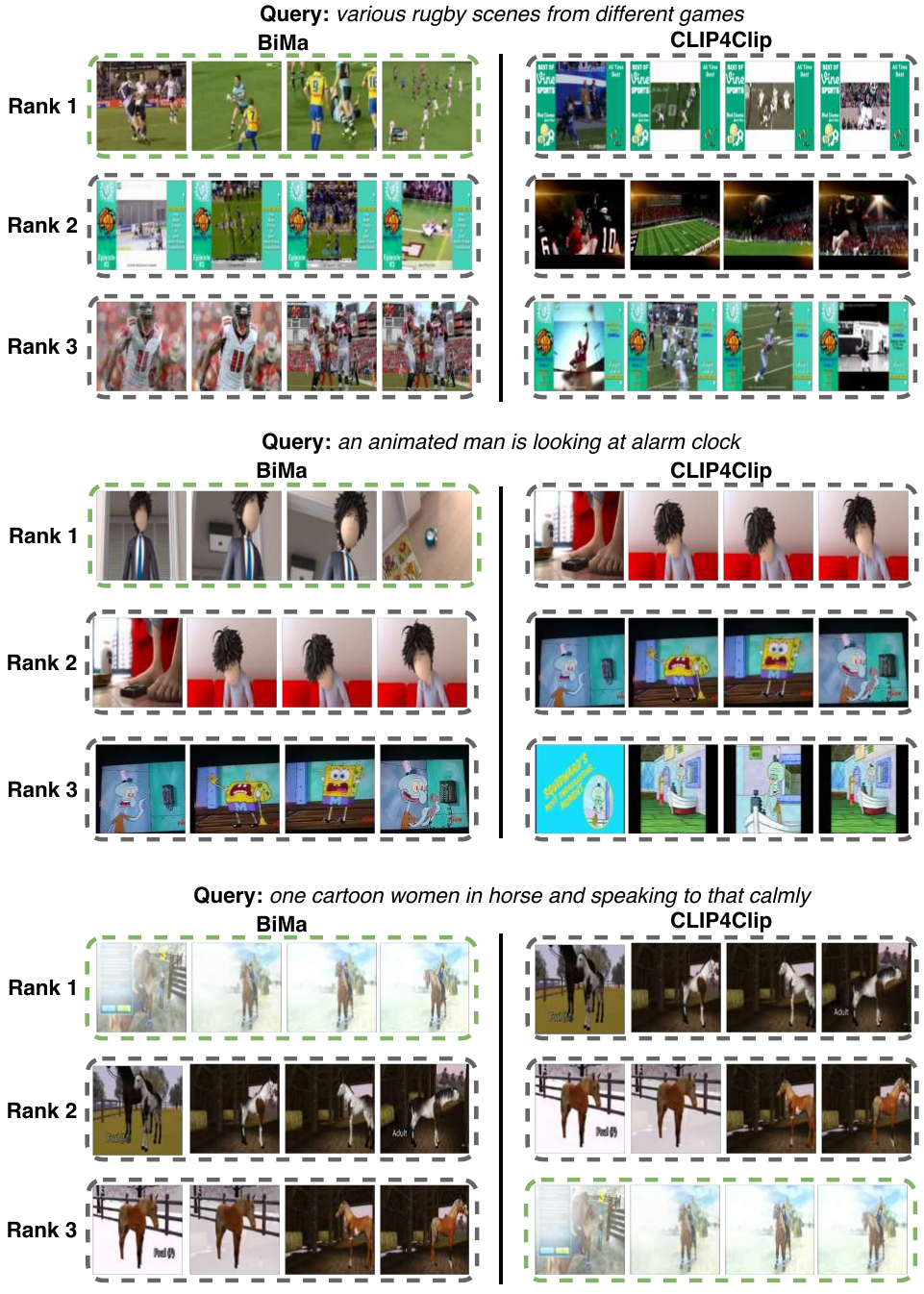} 
    \caption{\textbf{Text-to-video Results.} On the left are \texttt{BiMa}'s retrieval results and on the right are \texttt{CLIP4Clip}'s retrieval results. Ground truth videos are highlighted in the green box. These results demonstrate that our method can align the correlation between text and video effectively.}
    \label{fig:retrieval_2}
\end{figure*}

\begin{figure*}[!t]
    \centering
    \includegraphics[width=0.65\linewidth]{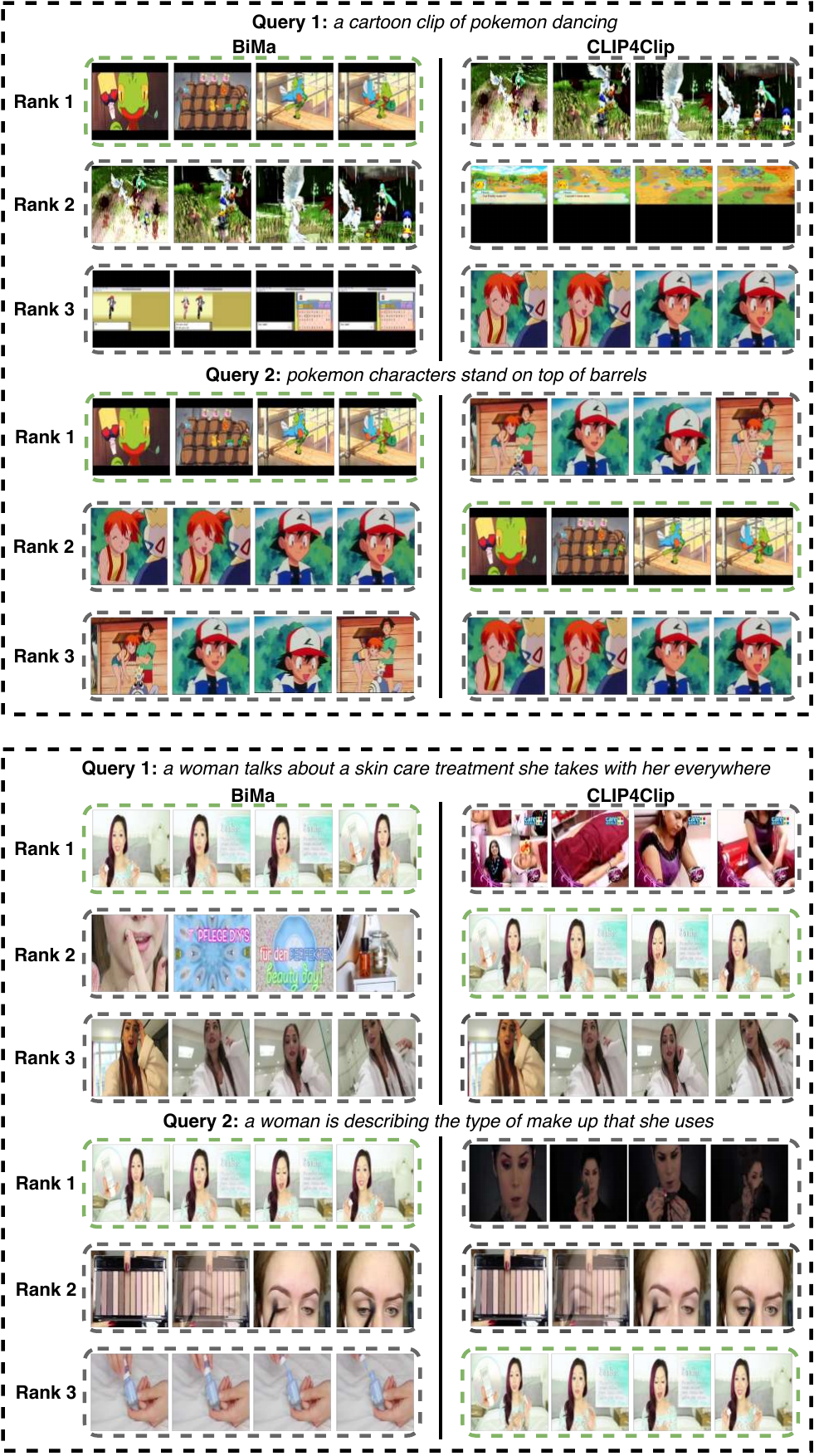} 
    \caption{\textbf{Different Queries Text-to-video Results.} We show retrieval results on two different queries of the same ground truth video. On the left are \texttt{BiMa}'s retrieval results and on the right are \texttt{CLIP4Clip}'s retrieval results. Ground truth videos are highlighted in the green box. These results demonstrate that our method can retrieve consistently despite different query content.}
    \label{fig:retrieval_3}
\end{figure*}

To further illustrate the debias capabilities of our \texttt{BiMa}, we conduct additional  qualitative comparisons between our \texttt{BiMa} and \texttt{CLIP4Clip} as follows:

\begin{itemize}
    \item Figs.~\ref{fig:word_cloud} and~\ref{fig:scene_elements_statistics} respectively show the word clouds and statistics on dataset-specific scene elements that we extracted based on our taxonomy dictionary. 
    \item Figs.~\ref{fig:attn_1},~\ref{fig:attn_2}, and~\ref{fig:attn_3} show the visual attention heatmaps generated by our \texttt{BiMa} model in comparison with \texttt{CLIP4Clip} on the T2V task. 
    \item Figs.~\ref{fig:retrieval_1} and~\ref{fig:retrieval_2} illustrate the robustness and efficacy of the proposed \texttt{BiMa}.
    \item Fig.~\ref{fig:retrieval_3} demonstrates how the same video can be retrieved using different input text queries.
\end{itemize}
\begin{table*}
    \setlength{\tabcolsep}{3pt}
    \centering
    \resizebox{0.97\linewidth}{!}{
    
    \begin{tabular}{@{}c|m{17cm}@{}}
    	\toprule[1.25pt]
                \textbf{Dataset}  & \textbf{Examples}\\ 
            \midrule
              & 1) A bulldozer removes dirt \\
        MSR-VTT & 2) An infomercial with a pharmeceutical company talking about an epilepsy drug pending approval from the FDA \\
                 ($\sim$43 characters & 3) Extreme violence scenes with people fighting with each other \\
                in a text) & 4) A woman is in front of a whiteboard talking about the numbers written on it \\
                & 5) A man is playing an instrument \\

 \midrule

 & 1) A lady is pouring raw strawberry juice into a bowl \\
MSVD & 2) A man is slicing the crust into a potato \\ 
 ($\sim$31 characters& 3) A man lifts three sunflowers \\ 
  in a text) & 4) A man is putting a pan into an oven \\
 & 5) A boy rides around in circles on a tricycle \\
\midrule

 & 1) The dish is covered in saffron and spices. \\
LSMDC & 2) He slaps SOMEONE again.  \\
($\sim$46 characters& 3) He and SOMEONE join forces to grab the cube, which's connected with several more wire. \\
 in a text) & 4) In the race, a rider falls. \\
 & 5) SOMEONE dashes to a clothes closet and ducks inside.
The cup spins across the floor. \\
\midrule

 & 1) A man is seen speaking to the camera and pans out into more men standing behind him. \\
ActivityNet & 2) A woman is seen kneeling down next to a man.  \\
($\sim$67 characters& 3) Several people are shown doing long boarding stunts in the parking lots of a college. \\
 in a text) & 4) People are sitting inside of a raft going over large bumps in the water. \\
 & 5) Several shots are shown of a man speaking to groups of people and leads into people wearing wet suits and walking. \\
                
\midrule
  \multirow{6}{*}{DiDeMo}  & 1) First time we see the dancers go down on one leg the men hit the ground with their sticks. They first start crouching and hitting the ground with the sticks \\
\multirow{7}{*}{in a text)} & 2) When the man puts his head down the guitar player is looking up. The guitarist is looking straight up. A man plays the guitar while looking up. The guitarist is looking straight up as he plays. \\
  
($\sim$147 characters  & 3) Red phone booth is visible a red phone booth is in the scene. A person walks in the middle of the camera. A red phone booth can be seen a red telephone booth is on the sidewalk. \\

   & 4) The camera moves back to the left to the tree. White square exits frame left the camera pans back the way it came. Square area lines with stones comes into view the fence comes into view. \\

& 5) Fog moves in toward the ice skater. a woman spins around several times very fast. A woman pirouettes as she comes near the camera. Woman spins more than 5 times in a row. \\

    \arrayrulecolor{black}
    \bottomrule[1.25pt]
    \end{tabular}
    }
    \caption{Five random examples of text descriptions in different datasets. For each dataset, we also report the median length of a text in that dataset on the left column.} 
    \label{tab:text_styles}

    \vspace{-0.25cm}
\end{table*}

\textbf{Attention Map Comparison}
To gain deeper insights into the bias mitigation capability of our \texttt{BiMa}, we conduct the attention map visualizations generated using \texttt{gradCAM}~\cite{selvaraju2017grad} between our \texttt{BiMa} and the baseline model \texttt{CLIP4Clip}, as illustrated in Figs.~\ref{fig:attn_1},~\ref{fig:attn_2}, and~\ref{fig:attn_3}. 
The attention maps reveal that our proposed \texttt{BiMa} effectively captures the essence of scene elements in the video. 
As shown in the second frame of Figs.~\ref{fig:attn_2} and \ref{fig:attn_3}, we observe that \texttt{BiMa} can focus on small-size actors (e.g., the singer and the horse) and disregard the distractive environment. Further, the attention map completely matches the top-3 scene entities, which also can be employed as explainability in further research.
These results demonstrate \texttt{BiMa} ability to capture crucial scene elements present in the video, thereby enabling successful video retrieval.

\textbf{Text-to-Video Retrieval}
In Figs.~\ref{fig:retrieval_1} and~\ref{fig:retrieval_2}, we provide Text-to-video (T2V) comparisons between our \texttt{BiMa} and the baseline model \texttt{CLIP4Clip}. Remarkably, our method consistently retrieves the ground truth video as the 1st rank video. These results demonstrate that our method can align video and text effectively.

Additionally, we present additional retrieval examples in Fig.~\ref{fig:retrieval_3}, utilizing various textual queries of the same ground truth video. This figure serves as a comparison between our \texttt{BiMa} and the baseline model \texttt{CLIP4Clip}. We observe that our proposed \texttt{BiMa} consistently retrieves the ground truth video across different textual queries, while \texttt{CLIP4Clip} often retrieves incorrect results.

\end{document}